\documentclass[journal]{IEEEtran}

\usepackage{xcolor} % Package to handle colors

\usepackage{cite} % Cites and compresses citation numbers

\usepackage[pdftex]{graphicx}
\graphicspath{{../figures/}}
\DeclareGraphicsExtensions{.pdf} %,.jpeg,.png}

\usepackage{amsmath}
\usepackage{amsthm}
\usepackage{amsfonts}
\usepackage{amssymb}

\usepackage{array} % Enhances array, tabular environments (strongly encouraged)

\usepackage[caption=false,font=footnotesize]{subfig}
\usepackage{stfloats} % Double column floats placed at the bottom
\usepackage{url} % Handling and breaking URLs

\input{auxFiles/mySymbol.sty}
\usepackage{needspace}

\newcommand{\myparagraph}[1]{\needspace{1\baselineskip}\medskip\noindent {\bf #1}}

\def\Tr{\mathsf{T}} % Transpose
\newcommand{\eg}{e.g.,\ }
\newcommand{\ie}{i.e.,\ }

\newtheorem{remark}{\hspace{0pt}\bf Remark}

\renewcommand{\blue}{}

\usepackage{booktabs}

\begin{document}
%%%%%%%%%%%%%%%%%%%%%%%%%%%%%%%%%%%%%%%%%%%%%%%%%%%%%%%%%%%%%%%%%%%%%%%%%%%%%%%%
%%%%                          DOCUMENT INFORMATION                          %%%%
%%%%%%%%%%%%%%%%%%%%%%%%%%%%%%%%%%%%%%%%%%%%%%%%%%%%%%%%%%%%%%%%%%%%%%%%%%%%%%%%

%%%%%%%%%%%%%%%%%%%%%%%%%%%%%%%%%%%%%%%%
%%%%          Paper Title           %%%%
%%%%%%%%%%%%%%%%%%%%%%%%%%%%%%%%%%%%%%%%

\title{Scalable Perception-Action-Communication Loops with Convolutional and Graph Neural Networks}

%%%%%%%%%%%%%%%%%%%%%%%%%%%%%%%%%%%%%%%%
%%%%            Authors             %%%%
%%%%%%%%%%%%%%%%%%%%%%%%%%%%%%%%%%%%%%%%

\author{Ting-Kuei~Hu, %
        Fer\hspace{0.015cm}nando~Gama, %
        Tianlong~Chen, %
        Wenqing~Zheng, %
        \\
        Zhangyang~Wang, %
        Alejandro~Ribeiro, %
        and~Brian~M.~Sadler% <-this % stops a space
    \thanks{T. Hu is with the Department of Computer Science and Engineering, Texas A\&M University, College Station, TX 77843 USA. Email: tkhu@tamu.edu.}% <-this % stops a space
    \thanks{F. Gama is with the Department of Computer and Electrical Engineering at Rice University, Houston, TX 77005 USA. Email: fgama@rice.edu}% <-this % stops a space
    \thanks{T. Chen, W. Zheng and Z. Wang are with the Department of Electrical and Computer Engineering, The University of Texas at Austin, Austin, TX 78712 USA. Email: \{tianlong.chen, w.zheng, atlaswang\}@utexas.edu.}% <-this % stops a space
    \thanks{A. Ribeiro is with the Department of Electrical and Systems Engineering, the University of Pennsylvania, Philadelphia, PA 19104 USA. Email: aribeiro@seas.upenn.edu.}% <-this % stops a space
    \thanks{B. Sadler is with the US Army Research Laboratory, Adelphi, MD 20783 USA. Email: brian.m.sadler6.civ@army.mil.}% <-this % stops a space
    \thanks{The work was in part supported by the ARL Distributed and Collaborative Intelligent Systems and Technology, Collaborative Research Alliance.}}% <-this % stops a space

%%%%%%%%%%%%%%%%%%%%%%%%%%%%%%%%%%%%%%%%
%%%%            Headers             %%%%
%%%%%%%%%%%%%%%%%%%%%%%%%%%%%%%%%%%%%%%%

\markboth{IEEE TRANSACTIONS ON SIGNAL AND INFORMATION PROCESSING OVER NETWORKS (SUBMITTED)}%
{Hu \MakeLowercase{\textit{et al.}}: Scalable PAC Loops with CNNs and GNNs}

%%%%%%%%%%%%%%%%%%%%%%%%%%%%%%%%%%%%%%%%
%%%%           MAKE TITLE           %%%%
%%%%%%%%%%%%%%%%%%%%%%%%%%%%%%%%%%%%%%%%

\maketitle

%%%%%%%%%%%%%%%%%%%%%%%%%%%%%%%%%%%%%%%%%%%%%%%%%%%%%%%%%%%%%%%%%%%%%%%%%%%%%%%%
%%%%                                ABSTRACT                                %%%%
%%%%%%%%%%%%%%%%%%%%%%%%%%%%%%%%%%%%%%%%%%%%%%%%%%%%%%%%%%%%%%%%%%%%%%%%%%%%%%%%

\begin{abstract}
In this paper, we present a perception-action-communication loop design using Vision-based Graph Aggregation and Inference (VGAI). This multi-agent decentralized learning-to-control framework maps raw visual observations to agent actions, aided by local communication among neighboring agents. Our framework is implemented by a cascade of a convolutional and a graph neural network (CNN / GNN), addressing agent-level visual perception and feature learning, as well as swarm-level communication, local information aggregation and agent action inference, respectively. By jointly training the CNN and GNN, image features and communication messages are learned in conjunction to better address the specific task. We use imitation learning to train the VGAI controller in an offline phase, relying on a centralized expert controller. This results in a learned VGAI controller that can be deployed in a distributed manner for online execution. Additionally, the controller exhibits good scaling properties, with training in smaller teams and application in larger teams. Through a multi-agent flocking application, we demonstrate that VGAI yields performance comparable to or better than other decentralized controllers, using only the visual input modality and without accessing precise location or motion state information.
% EDICS:
% SIPG-APP Applications < SIPG Signal and Information Processing over Graphs
\end{abstract}

%%%%%%%%%%%%%%%%%%%%%%%%%%%%%%%%%%%%%%%%%%%%%%%%%%%%%%%%%%%%%%%%%%%%%%%%%%%%%%%%
%%%%                                KEYWORDS                                %%%%
%%%%%%%%%%%%%%%%%%%%%%%%%%%%%%%%%%%%%%%%%%%%%%%%%%%%%%%%%%%%%%%%%%%%%%%%%%%%%%%%


% \fi
\IEEEpeerreviewmaketitle

%%%%%%%%%%%%%%%%%%%%%%%%%%%%%%%%%%%%%%%%%%%%%%%%%%%%%%%%%%%%%%%%%%%%%%%%%%%%%%%%
%%%%                                                                        %%%%
%%%%                              INTRODUCTION                              %%%%
%%%%                                                                        %%%%
%%%%%%%%%%%%%%%%%%%%%%%%%%%%%%%%%%%%%%%%%%%%%%%%%%%%%%%%%%%%%%%%%%%%%%%%%%%%%%%%

\section{Introduction} \label{sec:intro}

%!TEX root = 00-VGAI.tex

%%%%%%%%%%%%%%%%%%%%%%%%%%%%%%%%%%%%%%%%%%%%%%%%%%%%%%%%%%%%%%%%%%%%%%%%%%%%%%%%
%%%%                                                                        %%%%
%%%%                              INTRODUCTION                              %%%%
%%%%                                                                        %%%%
%%%%%%%%%%%%%%%%%%%%%%%%%%%%%%%%%%%%%%%%%%%%%%%%%%%%%%%%%%%%%%%%%%%%%%%%%%%%%%%%
%%%% sec:intro
%%%%%%%%%%%%%%

Large-scale aerial swarms  
of collaborating agents are under study
for wireless networking, disaster response, and military situational awareness, among many other applications. Many approaches rely on a centralized controller with global state information, such as using IMU sensor measurements, or global navigation satellite sensing 
\cite{Mellinger2011-SnapQuad, Kushleyev2013-Agile, Preiss2017-CrazySwarm, Vasarhelyi2014-Flocking, Vasarhelyi2018-FlockingConfined, Weinstein2018-VisionSwarm}. 
Centralized controllers assume access to global information at each time step, and implement some form of optimal global policy. Centralized control may be reasonably implemented for smaller swarm sizes, but does not scale efficiently to larger numbers. The centralized approach may lack robustness to time-varying networking quality or the failure of a leader or fusion node. As the number of agents grows  these issues become dominant, with increasing communication overhead and complexity, resulting in delays or channel deterioration.

Decentralized control addresses many of these issues by relying on local perception, and may include communications between agents. Recent work in perception-action-control (PAC) loops generally incorporates local sensing, state estimation, and communications with neighboring agents \cite{Paulos2019-PAC, Lee2020-PAC, Nguyen2020-PAC, Sun2020-PAC, Gama2021-ControlGNN}. Decentralized controllers have exhibited better properties in terms of scalability, robustness and communication resource allocation.

Robotic sensors such as IMU's have been incorporated into many decentralized controllers for robot swarm coordination \cite{Mellinger2011-SnapQuad, Kushleyev2013-Agile, Preiss2017-CrazySwarm, Weinstein2018-VisionSwarm}. However, advances in miniature vision sensors, computer vision, and deep learning allow access to unparalleled vision information density for autonomous systems that is yet to be fully exploited for the control of robot swarms \cite{Wu2019-ObjectUnmanned}. Visual inputs can capture a change of location or relative velocity of other agents in its field of view (potentially long-range, and covering more than one-hop networking neighbors), with virtually no delay when compared with wireless networking. Using visual information also leads to enhanced robustness with respect to network failures and adversarial compromise. These characteristics strongly motivate use of vision as an integral part of decentralized swarm control.

Nevertheless, developing a decentralized control system based on local visual observations raises  unique challenges. While direct onboard measurements of the agent state (velocity, location) are readily related to control actions, raw visual inputs are harder to interpret, and are typically more costly to process or communicate. An end-to-end mapping from raw visual input to end actions has been studied for small-scale swarms (9 agents),  
and collision avoiding leader-follower reactive control was demonstrated, without communications between agents \cite{Schilling2018-LearningVision}.
%albeit without any consideration of communication capabilities between agents, thus rendering this approach difficult to scale.

%%%%%%%%%%%%%%%%%%%%%%%%%%%%%%%%%%%%%%%%
%%%%             FIGURE             %%%%  fig:VGAI
%%%%%%%%%%%%%%%%%%%%%%%%%%%%%%%%%%%%%%%%
%%
\begin{figure*}[!t]
    \centering
    \includegraphics[width=0.9\linewidth]{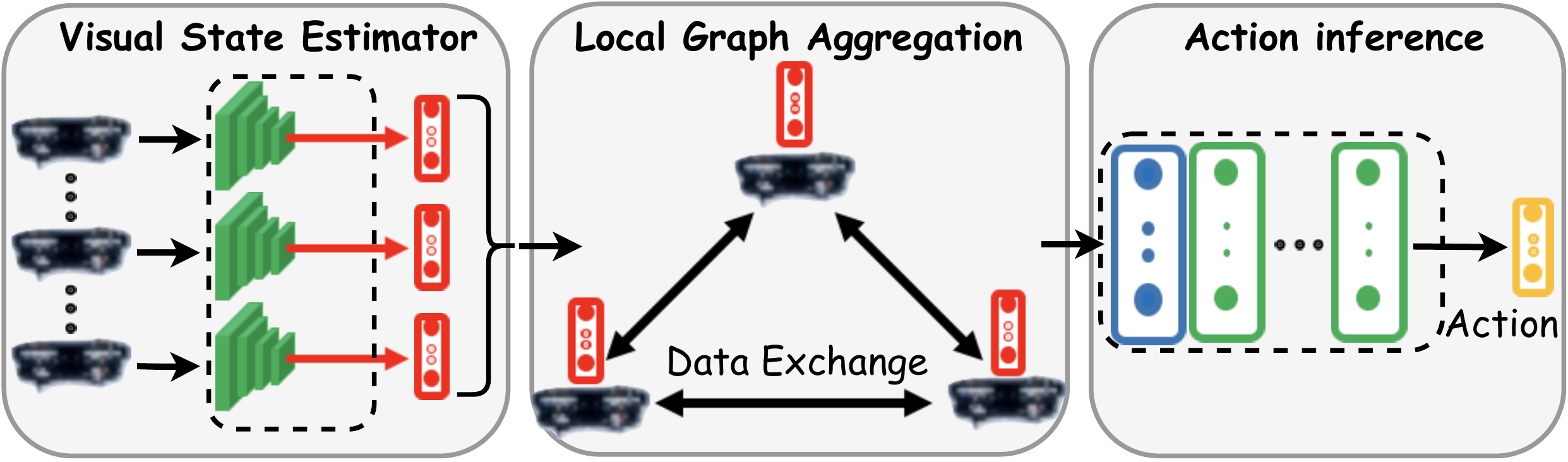}
    \caption{The proposed VGAI framework. Stage (i), each agent's raw visual observation is mapped into a compact local feature descriptor. Stage (ii), each agent communicates locally and aggregates its neighbors features. Stage (iii), each agent generates a control based on the aggregated features.}
    \label{fig:VGAI}
\end{figure*}
%%
%%%%          End of FIGURE         %%%%
%%%%%%%%%%%%%%%%%%%%%%%%%%%%%%%%%%%%%%%%

In this paper  we propose \textit{Vision-based Graph Aggregation and Inference} (VGAI), a decentralized learning-to-control framework that directly maps raw visual observations to agent actions, thus acting as a learnable PAC-loop controller. We combine visual processing and communications with neighbors, and rely on compact visual information representation to maintain relatively low bandwidth networking requirements. As illustrated in Fig. \ref{fig:VGAI}, VGAI consists of three stages: \textbf{(i)} visual state estimation; \textbf{(ii)} local graph aggregation; and \textbf{(iii)} action inference. Stages (i) and (iii)  are carried out by each agent individually, while Stage (ii) involves local sparse communication. VGAI is implemented by a cascade of a convolutional neural network (CNN) and a graph neural network (GNN), addressing Stage (i), and Stages (ii) and (iii), respectively.  Each agent has a CNN that maps visual input to a compact, local visual descriptor, for efficient transmission to neighbors. 

We leverage two recently proposed  graph neural network (GNN) \cite{Gama2020-GNNs, Ruiz2021-GNNs} learning frameworks for Stage (ii), namely Delayed Aggregation Graph Neural Network (DAGNN) \cite{Gama2019-Archit, Tolstaya2019-Flocking}, and Graph Recurrent Neural Network (GRNN) \cite{Ruiz2020-GRNN}.  Each agent fuses its neighbors  visual descriptors with its own,  and outputs a control step. The communication is local based on the graph topology (i.e., it requires, at most, repeated exchanges with the one-hop neighbors only), and the local connectivity patterns are dictated by the adopted network graph model. In our examples, we consider both Disk and K-nearest neighbor communication models, to explore the effect of different local connectivity assumptions.

VGAI employs a seamless integration of \emph{agent visual perception} and \emph{local communication}. Thanks to the latter, VGAI is able to scale up to medium- and large-sized swarms, and we present an example with 75 agents. We examine the proposed VGAI framework for a drone flocking application \cite{Tolstaya2019-Flocking}. Extensive experiments demonstrate that VGAI outperforms other competing decentralized controllers, and achieves comparable performance to the centralized controller that learns from global information. VGAI is also shown to generalize to several more challenging visual environments, including previously unobserved visual degradation, and complex visual backgrounds, with the aid of CNN pre-training.

This paper builds on and significantly extends our preliminary conference version of VGAI \cite{Hu2021-VGAI}. We have extended the GNN architecture to incorporate time memory by using graph recurrent neural networks. This captures information from multi-hop neighbors, and allows inference over time dependent states. We extend visual training and consider different challenging visibility environments, and visual generalization and pre-training. We also consider different networking connectivity assumptions. Extensive experiments for flocking drones illustrate the CNN and GNN learning architecture and show the value of combined learning with vision, networking, and control.

%%%%%%%%%%%%%%%%%%%%%%%%%%%%%%%%%%%%%%%%%%%%%%%%%%%%%%%%%%%%%%%%%%%%%%%%%%%%%%%%
%%%%                                                                        %%%%
%%%%                              RELATED WORK                              %%%%
%%%%                                                                        %%%%
%%%%%%%%%%%%%%%%%%%%%%%%%%%%%%%%%%%%%%%%%%%%%%%%%%%%%%%%%%%%%%%%%%%%%%%%%%%%%%%%

\section{Related Work} \label{sec:related}

%!TEX root = 00-VGAI.tex

%%%%%%%%%%%%%%%%%%%%%%%%%%%%%%%%%%%%%%%%%%%%%%%%%%%%%%%%%%%%%%%%%%%%%%%%%%%%%%%%
%%%%                                                                        %%%%
%%%%                              RELATED WORK                              %%%%
%%%%                                                                        %%%%
%%%%%%%%%%%%%%%%%%%%%%%%%%%%%%%%%%%%%%%%%%%%%%%%%%%%%%%%%%%%%%%%%%%%%%%%%%%%%%%%
%%%% sec:related
%%%%%%%%%%%%%%%%

We describe related work in three categories. Sec.~\ref{subsec:related:decentralizedFlocking} reviews literature on the decentralized control of a flock of drones. Sec.~\ref{subsec:related:visionSingle} summarizes recent advances in general vision-based drone control, and Sec.~\ref{subsec:related:visionFlocking} focuses on existing approaches that are both vision-based and decentralized.

%%%%%%%%%%%%%%%%%%%%%%%%%%%%%%%%%%%%%%%%%%%%%%%%%%%%%%%%%%%%%%%%%%%%%%%%%%%%%%%%
%%%%            SUBSECTION : Decentralized Flocking with Drones             %%%%
%%%%%%%%%%%%%%%%%%%%%%%%%%%%%%%%%%%%%%%%%%%%%%%%%%%%%%%%%%%%%%%%%%%%%%%%%%%%%%%%
%%%% subsec:related:decentralizedFlocking
%%%%%%%%%%%%

\subsection{Decentralized flocking with drones} \label{subsec:related:decentralizedFlocking}

Centralized controllers are able to access global information to decide on optimal control actions \cite{Reynolds1987-DistributedBehavior, Tanner2003-Stable}, but are not easily realized for large-scale ad hoc swarm deployments. Bio-inspired decentralized aerial swarm robotics address this through local controllers using local processing and neighbor information exchange \cite{DiCaro2013-EfficientSwarm, Floreano2015-FutureDrones}.  
However, it has long been known that finding optimal controllers in
these distributed settings is challenging, 
primarily due to the locality restriction of network communication \cite{Witsenhausen1968-Counterexample, Gama2021-DistributedLQR}.

%Recently, decentralized learning approaches have been successfully applied in various topics, such as wireless communications, tracking and navigations \cite{DecentralizedInference, giannakis2015decentralized, WeightedADMM, FastDecentralizedLearning, liang2020asynchronous, lin2021decentralized, CNN_Drone, DecentralizedReinforcement}.
Decentralized flocking algorithms design local controllers that incorporate local observations from neighbors \cite{Reynolds1987-DistributedBehavior, Tanner2003-Stable, Jadbabaie2003-Coordination}, and can be extended to include information exchanges from multi-hop neighbors \cite{Tolstaya2019-Flocking, Ruiz2020-GRNN}. These works typically assume the availability of precise location or motion, e.g., requiring access to global navigation satellite system (GNSS) positioning, although positioning may be imprecise or unavailable in many scenarios. 

%%%%%%%%%%%%%%%%%%%%%%%%%%%%%%%%%%%%%%%%%%%%%%%%%%%%%%%%%%%%%%%%%%%%%%%%%%%%%%%%
%%%%             SUBSECTION : Vision-based Single Drone Control             %%%%
%%%%%%%%%%%%%%%%%%%%%%%%%%%%%%%%%%%%%%%%%%%%%%%%%%%%%%%%%%%%%%%%%%%%%%%%%%%%%%%%
%%%% subsec:related:visionSingle
%%%%%%%%%%%%

\subsection{Vision-based single drone control} \label{subsec:related:visionSingle}

Imitation learning \cite{Ross2010-ImitationLearning, Hussein2017-ImitationLearning, Osa2018-ImitationLearning} has been commonly used in vision-based drone control design, especially for collision avoidance in complex environments.  An early work extracts image features and maps these to control input for avoiding trees while flying in a forest, and relies on expert training \cite{Ross2013-Monocular}. An approach to collision avoidance, called DroneNet \cite{Loquercio2018-DroNet}, uses supervised learning to train a CNN with visual input to predict collision probability as a function of steering angle. This is used to control steering angle, and forward velocity is modulated by the collision probability. Reinforcement learning using a neural network trained in a simulated environment can generalize to real-world navigation minimizing collisions \cite{Sadeghi2017-CAD2RL}. Other data-driven approaches \cite{Giusti2016-RobotForest, Gandhi2017-FlyCrash, Smolyanskiy2017-MAV} have also shown generality for flying in real-world environments. 
These methods offer enhanced single-agent operation and collision avoidance behavior.

%%%%%%%%%%%%%%%%%%%%%%%%%%%%%%%%%%%%%%%%%%%%%%%%%%%%%%%%%%%%%%%%%%%%%%%%%%%%%%%%
%%%%            SUBSECTION : Vision-based decentralized flocking            %%%%
%%%%%%%%%%%%%%%%%%%%%%%%%%%%%%%%%%%%%%%%%%%%%%%%%%%%%%%%%%%%%%%%%%%%%%%%%%%%%%%%
%%%% subsec:related:visionFlocking
%%%%%%%%%%%%

\subsection{Vision-based decentralized flocking} \label{subsec:related:visionFlocking}

Visual identification of neighbors can be enhanced by mounting unique visual markers on each agent, although this has practical issues \cite{Faigl2013-LowCost, Krajnik2014-PracticalLocalization}. In \cite{Schilling2018-LearningVision}, Schilling, et. al, developed a 
basic
%primitive 
decentralized vision-based flocking approach that generates velocity commands from raw camera images using a CNN. 
This reactive control does not use communications between agents. Collision avoidance was demonstrated in  leader-follower scenarios.
%However, the lack of spatial and temporal information exchanges between nearby agents make the learned policy difficult to be utilized in large scale multi-agent systems.

Our proposed approach is able to work on large swarms by exploiting local communication between nearby agents based on recent progress in graph neural networks \cite{Tolstaya2019-Flocking, Ruiz2020-GRNN}. Joint training of visual feature estimation and GNN-based local communications provides a scalable approach, and we demonstrate this with a large swarm of 75 nodes.

%%%%%%%%%%%%%%%%%%%%%%%%%%%%%%%%%%%%%%%%%%%%%%%%%%%%%%%%%%%%%%%%%%%%%%%%%%%%%%%%
%%%%                                                                        %%%%
%%%%                                FLOCKING                                %%%%
%%%%                                                                        %%%%
%%%%%%%%%%%%%%%%%%%%%%%%%%%%%%%%%%%%%%%%%%%%%%%%%%%%%%%%%%%%%%%%%%%%%%%%%%%%%%%%

\section{Flocking} \label{sec:flocking}

%!TEX root = 00-VGAI.tex

%%%%%%%%%%%%%%%%%%%%%%%%%%%%%%%%%%%%%%%%%%%%%%%%%%%%%%%%%%%%%%%%%%%%%%%%%%%%%%%%
%%%%                                                                        %%%%
%%%%                                FLOCKING                                %%%%
%%%%                                                                        %%%%
%%%%%%%%%%%%%%%%%%%%%%%%%%%%%%%%%%%%%%%%%%%%%%%%%%%%%%%%%%%%%%%%%%%%%%%%%%%%%%%%
%%%% sec:flocking
%%%%%%%%%%%%%%%%%

Consider a set of $N$ agents $\ccalV = \{1,\ldots,N\}$. At time $t \in \mbN_{0}$, each agent $i \in \ccalV$ is described by its position $\bbr_{i}(t) = [r_{i}^{x}(t), r_{i}^{y}(t)]^{\Tr} \in \reals^{2}$, velocity $\bbv_{i}(t) = [v_{i}^{x}(t),v_{i}^{y}(t)]^{\Tr} \in \reals^{2}$ and acceleration $\bbu_{i}(t) = [u_{i}^{x}(t),u_{i}^{y}(t)]^{\Tr} \in \reals^{2}$. Let $t$ be a discrete-time index representing consecutive time sampling instances with interval $T_{s}$. The evolution of the system is then given by
% eqn:discreteDynamics
\begin{equation} \label{eqn:discreteDynamics}
\begin{aligned}
\bbr_{i}(t+1) &= \bbu_{i}(t) T_{s}^{2}/2 + \bbv_{i}(t) T_{s} + \bbr_{i}(t) \\
\bbv_{i}(t+1) &= \bbu_{i}(t) T_{s} + \bbv_{i}(t)
\end{aligned}
\end{equation}
for $t=0,1,2,\ldots$, which implies that each acceleration $\bbu_{i}(t)$ is held constant in the interval $[t T_{s},(t+1)T_{s})$. We further assume that transitions between $\bbu_{i}(t)$ and $\bbu_{i}(t+1)$ occur instantaneously.

To develop the VGAI framework, we consider multi-agent flocking. The objective of flocking is to coordinate the velocities $\bbv_{i}(t)$ of all agents to be the same
% eqn:flockingObjective
\begin{equation} \label{eqn:flockingObjective}
\min_{\substack{\bbu_{i}(t)\\ i=1,\ldots,N\\t \geq 0}} \frac{1}{N} \sum_{t} \sum_{i=1}^{N} \Big\| \bbv_{i}(t) - \frac{1}{N} \sum_{j=1}^{N} \bbv_{j}(t) \Big\|^{2}
\end{equation}
subject to the constraints enforced by the system dynamics \eqref{eqn:discreteDynamics}. A solution that avoids collisions is given by accelerations $\bbu_{i}^{\ast}(t)$, computed as \cite{Tanner2004-Flocking}
% eqn:optimalSolution
\begin{equation} \label{eqn:optimalSolution}
\bbu_{i}^{\ast}(t) = -\sum_{j=1}^{N} \Big( \bbv_{i}(t)-\bbv_{j}(t) \Big) - \sum_{j=1}^{N} \nabla_{\bbr_{i}(t)} U \Big( \bbr_{i}(t),\bbr_{j}(t) \Big) .
\end{equation}
Here
% eqn:collisionAvoidance
\begin{align}
& U(\bbr_{i}(t),\bbr_{j}(t)) 
\label{eqn:collisionAvoidance}\\
& =
\begin{cases} 
1/\|\bbr_{ij}(t)\|^{2} - \log(\|\bbr_{ij}(t)\|^{2}) & \text{if } \|\bbr_{ij}(t)\| \leq \rho \\
1/\rho^{2} - \log(\rho^{2}) & \text{otherwise}
\end{cases} \nonumber
\end{align}
is a collision avoidance potential, with $\bbr_{ij}(t) = \bbr_{i}(t) - \bbr_{j}(t)$ and $\rho$ is the value of the minimum distance allowed between agents. It is evident that, in computing the optimal solution \eqref{eqn:optimalSolution}, each agent $i$ requires knowledge of the velocities of all other agents in the network. Thus, the solution $\bbu^{\ast}(t)$ in \eqref{eqn:optimalSolution} is a \emph{centralized} controller. Note that we present the solution $\bbu^{\ast}(t)$ $\in \reals^{2}$ for 2D flocking problems where we assume that all agents work on the same plane. The solution $\bbu^{\ast}(t)$ can be extended for 3D flocking problems if we assume that its position $\bbr_{i}(t) = [r_{i}^{x}(t), r_{i}^{y}(t), r_{i}^{z}(t)]^{\Tr} \in \reals^{3}$, velocity $\bbv_{i}(t) = [v_{i}^{x}(t),v_{i}^{y}(t),v_{i}^{z}(t]^{\Tr} \in \reals^{3}$ and acceleration $\bbu_{i}(t) = [u_{i}^{x}(t),u_{i}^{y}(t),u_{i}^{z}(t)]^{\Tr} \in \reals^{3}$. In our experiments, for simplicity, we assume all agents fly at the same height, and use 2D vectors for the position and velocity (see also Remark 1 below).

Our objective, in contrast, is to obtain a \emph{decentralized} solution that can be computed only with information perceived by each agent, in combination with information relayed by neighboring agents. We describe the communication network by means of a succession of graphs $\ccalG(t) = \{\ccalV, \ccalE(t)\}$ where $\ccalV$ is the set of agents, and $\ccalE(t) \subseteq \ccalV \times \ccalV$ is the set of edges. The communication link $(i,j) \in \ccalE(t)$ allows for exchange of information between nodes $i$ and $j$ at time $t$. The existence of link $(i,j)$ can be assumed based on physical range or other aspects of the communications technology model. We denote by $\ccalN_{i}(t) = \{j \in \ccalV : (j,i) \in \ccalE(t)\}$ the set of all agents that can communicate with node $i$ at time $t$.

\begin{remark}[Three-dimensional environment] \normalfont
The current flocking description is based on a two-dimensional model [cf. \eqref{eqn:discreteDynamics}]. This aligns with our numerical experiments using the Microsoft\textregistered Airsim simulation environment, that only allows for four cameras on each drone. These capture front, left, right and back views, only providing enough information for a two-dimensional environment. The current framework, however, can be extended to three-dimensional scenarios. To do so, we would need to (1) utilize an expert controller that handles three-dimensional centralized flocking actions; (2) provide sufficient cameras on each drone to capture a full three-dimensional panoramic view; and (3) train a three-dimensional VGAI by replacing the two-dimensional inputs and ground truth with appropriate three-dimensional versions (Section~\ref{sec:visual}). The GNN part of the system would remain the same (Section~\ref{sec:gnn}).
\end{remark}

% A \emph{local heuristic} decentralized solution is to compute \eqref{eqn:optimalSolution} using information from each agents neighborhood \cite{Tolstaya2019-Flocking}, resulting in
% % eqn:localHeuristic
% \begin{align} \label{eqn:localHeuristic}
% \tbu_{i}(t) = & -\sum_{j\in \ccalN_{i}(t)} \Big( \bbv_{i}(t)-\bbv_{j}(t) \Big) \\
% & \quad - \sum_{j\in \ccalN_{i}(t)} \nabla_{\bbr_{i}(t)} U \Big( \bbr_{i}(t),\bbr_{j}(t) \Big). \nonumber
% \end{align}
% %
% This local heuristic decentralized controller assumes that the neighboring velocities, as well as the relative positions can be perfectly transmitted (the relative positions are required to compute $\nabla_{\bbr_{i}(t)}U$ as indicated in \cite[eq. (12)]{Tolstaya2019-Flocking}). We can potentially improve on this solution by incorporating information from neighbors that are farther away in the graph topology. However, under our assumptions about nearest neighbor communications, the information has to be relayed across agents while incurring networking delay. In what follows, we propose to \emph{learn} a decentralized solution that incorporates delayed information from multi-hop neighbors.

%%%%%%%%%%%%%%%%%%%%%%%%%%%%%%%%%%%%%%%%%%%%%%%%%%%%%%%%%%%%%%%%%%%%%%%%%%%%%%%%
%%%%                                                                        %%%%
%%%%                       VISUAL FEATURE EXTRACTION                        %%%%
%%%%                                                                        %%%%
%%%%%%%%%%%%%%%%%%%%%%%%%%%%%%%%%%%%%%%%%%%%%%%%%%%%%%%%%%%%%%%%%%%%%%%%%%%%%%%%

\section{Visual Feature Extraction} \label{sec:visual}

%!TEX root = 00-VGAI.tex

%%%%%%%%%%%%%%%%%%%%%%%%%%%%%%%%%%%%%%%%%%%%%%%%%%%%%%%%%%%%%%%%%%%%%%%%%%%%%%%%
%%%%                                                                        %%%%
%%%%                       VISUAL FEATURE EXTRACTION                        %%%%
%%%%                                                                        %%%%
%%%%%%%%%%%%%%%%%%%%%%%%%%%%%%%%%%%%%%%%%%%%%%%%%%%%%%%%%%%%%%%%%%%%%%%%%%%%%%%%
%%%% sec:visual
%%%%%%%%%%%%%%%

\begin{figure*}[!t]
    \centering
    \includegraphics[width=1.0\linewidth]{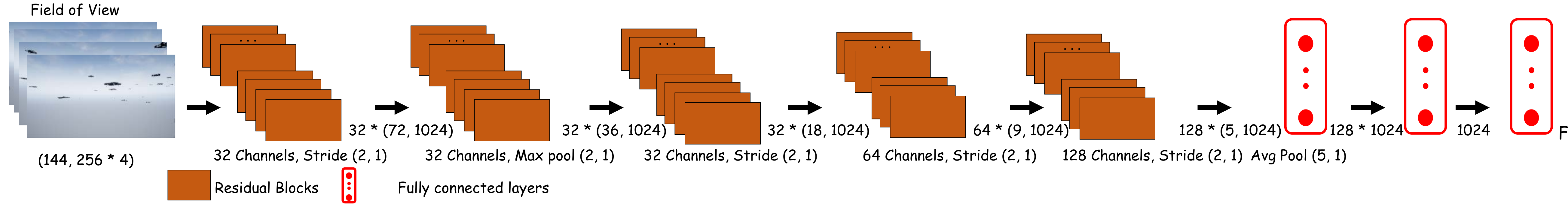}
    % \vspace{-1em}
    \caption{The architecture of the visual state estimator $\text{CNN}_{\bbPsi}(\cdot)$, consisting of five residual blocks and two fully-connected layers. The last layer's output, an $F$-dimensional vector, is the extracted visual state feature. For simplicity in this paper, the agents are assumed to be roughly at the same altitude, so (average) pooling is along the vertical axis while keeping the horizontal resolution intact.}
    \label{fig:uni_droNet}
\end{figure*}

In this section we consider the visual processing and feature extraction needed for VGAI. As shown in Figure~\ref{fig:uni_droNet}, the aim of the visual state estimator is to extract compact features $\bbX(t)$ from the raw visual observation $\bbH(t)$ (a collection of raw images) of local agents, that can indicate its motion state. The compact features can be further aggregated with neighboring agents information (described in the next section) for deciding the next control action. The visual state feature $\bbX(t)$ is obtained from a CNN denoted as $\text{CNN}_{\bbPsi}(\cdot)$, and the mapping of $\bbH(t)$ to $\bbX(t)$ is given by, 
% eq:CNN
\begin{equation} \label{eqn:CNN}
    \bbx_{i}(t) = \text{CNN}_{\bbPsi}\big( \bbh_{i}(t) \big),
\end{equation}
where $\bbPsi$ represents the set of learnable weights in all layers.  Here, $\bbx_{i}(t)$ and $\bbh_{i}(t)$ indicate the $i$th row of matrix $\bbX(t)$ and $\bbH(t)$, respectively. To aid in the scalability of the VGAI controller, we adopt a weight-sharing scheme by using the same mapping $\text{CCN}_{\Psi}$ for all agents, so that \eqref{eqn:CNN} is carried out row-wise and is therefore executed individually by each agent. 

We adopt DroNet \cite{Loquercio2018-DroNet} as the backbone architecture of $\text{CNN}_{\bbPsi}$, which consists of $10$ fast convolutions ($5$ residual blocks) and $2$ fully-connected layers as illustrated in Figure~\ref{fig:uni_droNet}. We note that it is desirable to keep the output feature dimension $F$ small for efficient transmission; an ablation study can be found in Section \ref{sec:sims}.

%%%%%%%%%%%%%%%%%%%%%%%%%%%%%%%%%%%%%%%%
%%%%             FIGURE             %%%%  fig:VGAI
%%%%%%%%%%%%%%%%%%%%%%%%%%%%%%%%%%%%%%%%
%%
%%
%%%%          End of FIGURE         %%%%
%%%%%%%%%%%%%%%%%%%%%%%%%%%%%%%%%%%%%%%%

%%%%%%%%%%%%%%%%%%%%%%%%%%%%%%%%%%%%%%%%%%%%%%%%%%%%%%%%%%%%%%%%%%%%%%%%%%%%%%%%
%%%%                                                                        %%%%
%%%%                       DECENTRALIZED CONTROLLERS                        %%%%
%%%%                                                                        %%%%
%%%%%%%%%%%%%%%%%%%%%%%%%%%%%%%%%%%%%%%%%%%%%%%%%%%%%%%%%%%%%%%%%%%%%%%%%%%%%%%%

\section{Decentralized Controllers} \label{sec:gnn}

%!TEX root = 00-VGAI.tex

%%%%%%%%%%%%%%%%%%%%%%%%%%%%%%%%%%%%%%%%%%%%%%%%%%%%%%%%%%%%%%%%%%%%%%%%%%%%%%%%
%%%%                                                                        %%%%
%%%%                       DECENTRALIZED CONTROLLERS                        %%%%
%%%%                                                                        %%%%
%%%%%%%%%%%%%%%%%%%%%%%%%%%%%%%%%%%%%%%%%%%%%%%%%%%%%%%%%%%%%%%%%%%%%%%%%%%%%%%%
%%%% sec:gnn
%%%%%%%%%%%%

Once we have extracted visual feature information [cf. Section~\ref{sec:visual}], we proceed to communicate this state with neighboring agents. The agents then need to learn a decentralized controller that is capable of inferring a suitable control action from their own state and those transmitted by immediate neighbors.

Let $\bbx_{i}(t) \in \reals^{F}$ be the state of agent $i$ at time $t$, described by an $F$-dimensional vector of \emph{features}. Denote by $\bbX(t) \in \reals^{N \times F}$ the row-wise collection of the state of all agents, given by
% eqn:stateMatrix
\begin{equation} \label{eqn:stateMatrix}
    \bbX(t) = 
        \begin{bmatrix}
            \bbx_{1}^{\Tr}(t) \\ \vdots \\ \bbx_{N}^{\Tr}(t)
        \end{bmatrix}.
\end{equation}
Note that $\bbX(t)$ represents the compact visual state features extracted from raw observations $\bbH(t)$ in \eqref{eqn:aggregationSequence}. To describe the communication between agents, we define the graph shift operator (GSO) matrix $\bbS(t) \in \reals^{N \times N}$ 
that reflects the sparsity of the graph, i.e., $[\bbS(t)]_{ij} = s_{ij}(t)$ is nonzero if and only if $(j,i) \in \ccalE(t)$. Examples of GSO used in the literature are the adjacency matrix \cite{Sandryhaila2013-DSPG}, the Laplacian matrix \cite{Shuman2013-SPG}, or respective normalizations \cite{Ortega2018-GSP}. Due to the sparsity of the GSO $\bbS(t)$, right-multiplication of $\bbS(t)$ with $\bbX(t)$ can be computed 
by means of local exchanges with neighboring nodes, yielding
% eqn:graphShift
\begin{equation} \label{eqn:graphShift}
    [\bbS(t) \bbX(t)]_{if} = \sum_{j \in \ccalN_{i}(t)} s_{ij}(t) [\bbx_{j}(t)]_{f}
\end{equation}
for each feature $f=1,\ldots,F$. In essence, multiplication \eqref{eqn:graphShift} updates the state at each agent using a linear combination of the states of neighboring agents. It is important to note that \eqref{eqn:graphShift} is a convenient mathematical description of the communication between agents, but that, in practice, there is no matrix multiplication involved, as $s_{ij}(t)$ represents the channel state of the link $(j,i)$ and therefore \eqref{eqn:graphShift} simply carries out a linear combination of the information transmitted by neighboring agents. As a matter of fact, each agent need not know the matrix $\bbS(t)$ nor the topology of the graph.

We propose two approaches to process the information obtained from neighbor exchanges: delayed-aggregation graph neural networks (Section~\ref{subsec:DAGNN}) and graph recurrent neural networks (Section~\ref{subsec:GRNN}), described next.

%%%%%%%%%%%%%%%%%%%%%%%%%%%%%%%%%%%%%%%%%%%%%%%%%%%%%%%%%%%%%%%%%%%%%%%%%%%%%%%%
%%%%         SUBSECTION : Delayed-Aggregation Graph Neural Networks         %%%%
%%%%%%%%%%%%%%%%%%%%%%%%%%%%%%%%%%%%%%%%%%%%%%%%%%%%%%%%%%%%%%%%%%%%%%%%%%%%%%%%
%%%% subsec:DAGNN
%%%%%%%%%%%%

\subsection{Delayed-aggregation graph neural networks} \label{subsec:DAGNN}

We build the \emph{aggregation sequence} \cite{Gama2019-Archit}, gathering information from multi-hop neighbors by means of $(K-1)$ repeated exchanges with one-hop neighbors. Thus, if $K=1$, then there are no information exchanges involved, if $K=2$ only one communication exchange with one-hop neighbors is carried out, if $K=3$ two communication exchanges are carried out, and so on. Note that if $K \ge 3$, then information beyond the one-hop neighbors is being relayed to the agent, albeit at the expense of a communication delay. Each agent then aggregates the received information, resulting in the \emph{aggregation sequence} $\bbZ^{d}(t)$ given by
% eqn:aggregationSequence
\begin{equation} \label{eqn:aggregationSequence}
\begin{aligned}
\bbZ^{d}(t) = \big[ 
& \bbX(t), \\
& \bbS(t) \bbX(t-1), \\
& \bbS(t) \bbS(t-1) \bbX(t-2), \\
& \ldots, \\
& \bbS(t) \cdots \bbS(t-(K-2)) \bbX(t-(K-1))
\big].
\end{aligned}
\end{equation}
The aggregation sequence $\bbZ^{d}(t)$ is an $N \times KF$ matrix, where each $N \times F$ block $\bbZ^{d}_{k}(t)$ represents the delayed aggregation of the state information at  $k$-hop neighbors. We denote $\bbz_{i}^{d}(t) \in \reals^{FK}$ to be the $i$th row of matrix $\bbZ^{d}(t)$, which represents the information gathered at node $i$ through $(K-1)$ communication exchanges with one-hop neighbors.

The information collected at a particular node depends on the graph topology. This is placed in vector $\bbz_{i}^{d}(t)$ and input to a neural network that maps the local information into a control action,
% eqn:NNlayer
\begin{equation} \label{eqn:NNlayer}
    \bbz_{\ell} = \sigma_{\ell} \big(\bbtheta_{\ell} \bbz_{\ell-1}^{d} \big) \ , \ \bbz_{0} = \bbz_{i}^{d}(t) \ , \ \bbu_{i}(t) = \bbz_{L}
\end{equation}
where $\bbz_{\ell} \in \reals^{F_{\ell}}$ represents the output of layer $\ell$, $\sigma_{\ell}$ is a pointwise nonlinearity (activation function), and $\bbtheta_{\ell} \in \reals^{F_{\ell} \times F_{\ell-1}}$ are the learnable parameters. The input to the neural network is the aggregation sequence $\bbz^{d}_{i}(t)$ at node $v_{i}$ [cf. \eqref{eqn:aggregationSequence}], with $F_{0} = FK$. We collect the resulting action as the output of the last layer $\bbu_{i}(t) = \bbz_{L}$ and thus $F_{L}$ represents the dimension of the control action to be taken. We compactly describe the neural network as
% eqn:NN
\begin{equation} \label{eqn:NN}
    \hbu_{i}(t) = \text{NN}_{\bbTheta}\big(\bbz_{i}^{d}(t) \big)
\end{equation}
where $\bbTheta = \{\bbtheta_{\ell}, \ell=1,\ldots,L\}$ are the 
learnable parameters of each layer.

We refer to this approach as a delayed-aggregation graph neural network (DAGNN). Several important observations are in order. First, the neural network parameters $\bbTheta$ do not depend on the specific node $i$, nor on the specific time-index $t$. This is a weight-sharing scheme that allows for scalability (i.e., once trained, it can be deployed on any number of agents), and prevents overfitting (i.e., it avoids growing the number of parameters with the number of agents). Second, since the aggregation sequence has already incorporated the graph information [cf. \eqref{eqn:aggregationSequence}], applying a conventional feed forward neural network to $\bbz_{i}(t)$ is already taking into account the underlying graph support \cite{Gama2019-Archit}. Third, the resulting architecture is entirely \emph{local} in the sense that, at test time, it can be implemented by means of repeated communication exchanges with one-hop neighboring nodes only. This is seen in \eqref{eqn:aggregationSequence}, which states that each node receives messages from their neighbors, processes them, and stores them in the corresponding row of $\bbZ^{d}(t)$. For each of the $K-1$ communication exchanges, each agent adds the messages obtained from immediate neighbors and stores it in the aggregation sequence \eqref{eqn:aggregationSequence}.

%%%%%%%%%%%%%%%%%%%%%%%%%%%%%%%%%%%%%%%%%%%%%%%%%%%%%%%%%%%%%%%%%%%%%%%%%%%%%%%%
%%%%              SUBSECTION : Graph Recurrent Neural Network               %%%%
%%%%%%%%%%%%%%%%%%%%%%%%%%%%%%%%%%%%%%%%%%%%%%%%%%%%%%%%%%%%%%%%%%%%%%%%%%%%%%%%
%%%% subsec:GRNN
%%%%%%%%%%%%

\subsection{Graph recurrent neural network} 
\label{subsec:GRNN}

As an alternative to the DAGNN described previously, 
we can capture the temporal dependencies of the states by means of a graph recurrent neural network (GRNN) \cite{Ruiz2020-GRNN}. To do this, we learn a \emph{hidden state} $\bbZ^{r}(t)$, which is a graph signal obtained by means of a nonlinear function that takes the current data and the previous hidden state as inputs, and outputs the updated hidden state. In particular, we choose a nonlinear function obtained by the cascade of graph convolutional filters and pointwise nonlinearities \cite{Ruiz2020-GRNN}.

A time-delayed graph convolutional filter is defined as
\begin{equation} \label{eqn:graph-conv}
    \bbA(\bbX;\bbS) = \sum_{k=0}^{K - 1} \bbS(t) \cdots \bbS(t-(k-1)) \bbX(t-k) \bbA_{k},
\end{equation}
where $\{\bbA_{k} \in \reals^{F \times G}\}$ is the set of $K$ filter coefficients which are learned from data and are used to determine the importance score of the information located in each $k$-hop neighborhood. A graph filter is capable of mapping agent states of dimension $F$ into agent states of dimension $G$. Note that the time-delay nature of the graph convolutional filter implies that each agent has access to its current information at time $t$, given by $\bbX(t)$ --the term for $k=0$--, the unit delayed information of their one-hop neighbor, given by $\bbS(t)\bbX(t-1)$ --the term for $k=1$--, the two-units delayed information of their two-hop neighbors, given by $\bbS(t)\bbS(t-1)\bbX(t-2)$ --the term for $k=2$--, and so on. Since $\bbS(t)$ respects the sparsity of the graph for all $t$, then $\bbS(t)\bbS(t-1)\cdots\bbX(t-k)$ represents a distributed operation that describes $k$ communication exchanges with one-hop neighbors, accounting for the corresponding delays. In practical terms, this implies that the nodes do not need to have access to the matrices $\{\bbS(t)\}$ since they only need to be able to communicate with their immediate neighbors and carry out a linear combination of the values received [cf. \eqref{eqn:graphShift}]. In this sense, $\bbS(t)\cdots\bbS(t-(k-1))\bbX(t-k)$ is a convenient mathematical formulation of the graph filter, but it does not represent the actual computational implementation (i.e., no matrix multiplications are involved).

The hidden state $\bbZ^{r}(t) \in \reals^{N \times H}$ is obtained from the transmission of the latest state $\bbX(t) \in \reals^{N \times F}$ across the network and from transmission of the previous hidden state $\bbZ^{r}(t-1) \in \reals^{N \times H}$ by means of graph convolutions followed by a pointwise nonlinearity $\sigma$ \cite{Ruiz2020-GRNN}
\begin{equation} \label{eqn:GRNN}
    \bbZ^{r}(t) = \sigma\big(\bbA(\bbX(t);\bbS) + \bbB(\bbZ^{r}(t-1);\bbS) \big).
\end{equation}
Recall that the filtering operations $\bbA$ and $\bbB$ in \eqref{eqn:GRNN} involve time-delays corresponding to the number of filter taps used [cf. \eqref{eqn:graph-conv}].
The filter $\bbA(\cdot;\bbS)$ maps the agent state of dimension $F$ into another state of dimension $H$, compatible with the dimension of the hidden state $\bbZ^{r}(t)$. The filter coefficients of $\bbA(\cdot;\bbS)$ and $\bbB(\cdot;\bbS)$ are different and they are both learned at training time. The hidden state $\bbZ^{r}(t) \in \reals^{N \times H}$ is also a graph signal, so essentially each node is learning its own hidden state by exchanging information with its neighbors. The hidden state $\bbZ^{r}(t)$ is intended to keep track of relevant information as each $\bbZ^{r}(t)$ is updated with the previous state $\bbZ^{r}(t-1)$ and with the new information in the signal $\bbX(t)$.

The hidden state captures information across the temporal dependencies of the state. Furthermore, since we are learning it from data, it captures the most relevant information for the specific task at hand. Then we can proceed to map the hidden state into a corresponding control action by means of another nonlinear mapping consisting of a graph convolution $\bbC (\cdot;\bbS)$ and another nonlinearity $\sigma_{o}$
% eqn:GRNNoutput
\begin{equation} \label{eqn:GRNNoutput}
    \bbU(t) = \sigma_{o} \big( \bbC( \bbZ^{r}(t) ; \bbS\big).
\end{equation}
Note that the graph filter $\bbC(\cdot;\bbS)$ maps the hidden state of dimension $H$ into the control action of dimension $G$. Jointly training the entire architecture, \eqref{eqn:GRNN}-\eqref{eqn:GRNNoutput}, results in learning the hidden state that captures both graph and temporal dependencies needed for decentralized control.

%%%%%%%%%%%%%%%%%%%%%%%%%%%%%%%%%%%%%%%%%%%%%%%%%%%%%%%%%%%%%%%%%%%%%%%%%%%%%%%%
%%%%            SUBSECTION : Training through imitation learning            %%%%
%%%%%%%%%%%%%%%%%%%%%%%%%%%%%%%%%%%%%%%%%%%%%%%%%%%%%%%%%%%%%%%%%%%%%%%%%%%%%%%%
%%%% subsec:imitation
%%%%%%%%%%%%

\subsection{Training through imitation learning} \label{subsec:imitation}

Both GNN architectures can be effectively trained by means of imitation learning \cite{Ross2010-ImitationLearning, Hussein2017-ImitationLearning, Osa2018-ImitationLearning}.
Consider training with \eqref{eqn:NN} for example.
We assume the availability of a training set consisting of trajectories obtained by employing some expert (usually centralized) controller, and our goal is to train the decentralized VGAI controller to imitate it. Here, we use \eqref{eqn:optimalSolution} in the case of flocking.
Note that while the centralized controller relies on global information, the resulting learned VGAI controller uses only local information. The training set is comprised of trajectories $\ccalT = \{(\bbX(t), \bbU^{\ast}(t))_{t}\}$, where $\bbX(t)$ is the collection of states \eqref{eqn:stateMatrix} and $\bbU^{\ast}(t) \in \reals^{N \times 2}$ is the collection of expert actions for each agent,
% eqn:actionMatrix
\begin{equation} \label{eqn:actionMatrix}
    \bbU^{\ast}(t) = \begin{bmatrix} \bbu_{1}^{\ast}(t)^{\Tr} \\ \vdots \\ \bbu_{N}^{\ast}(t)^{\Tr}
\end{bmatrix},
\end{equation}
where $\bbu_{i}^{\ast}(t) \in \reals^{2}$ is the action of agent $i$ at time $t$, given by the controller \eqref{eqn:optimalSolution} in the case of flocking. Then, the decentralized controller parameters can be found from solving
% eqn:imitationLearning
\begin{equation} \label{eqn:imitationLearning}
    \bbTheta^{\ast} = \argmin_{\bbTheta} \sum_{\ccalT} \sum_{i=1}^{N} \| \hbu_{i}(t) - \bbu_{i}^{\ast}(t) \|
\end{equation}
with $\hbu_{i}(t)$ being the VGAI controller. Recall from Section~\ref{sec:visual} that the local state $\bbx_{i}(t)$ is the output of the visual state estimator $\text{CNN}_{\bbPsi}(\cdot)$, and therefore solving \eqref{eqn:imitationLearning} is jointly learning the visual state estimator as well as the communication scheme.

\begin{figure*}[!t]
    \centering
    \subfloat[DAGNN\label{arch:DAGNN}]{
        \centering
        \includegraphics[height=0.29\textheight]{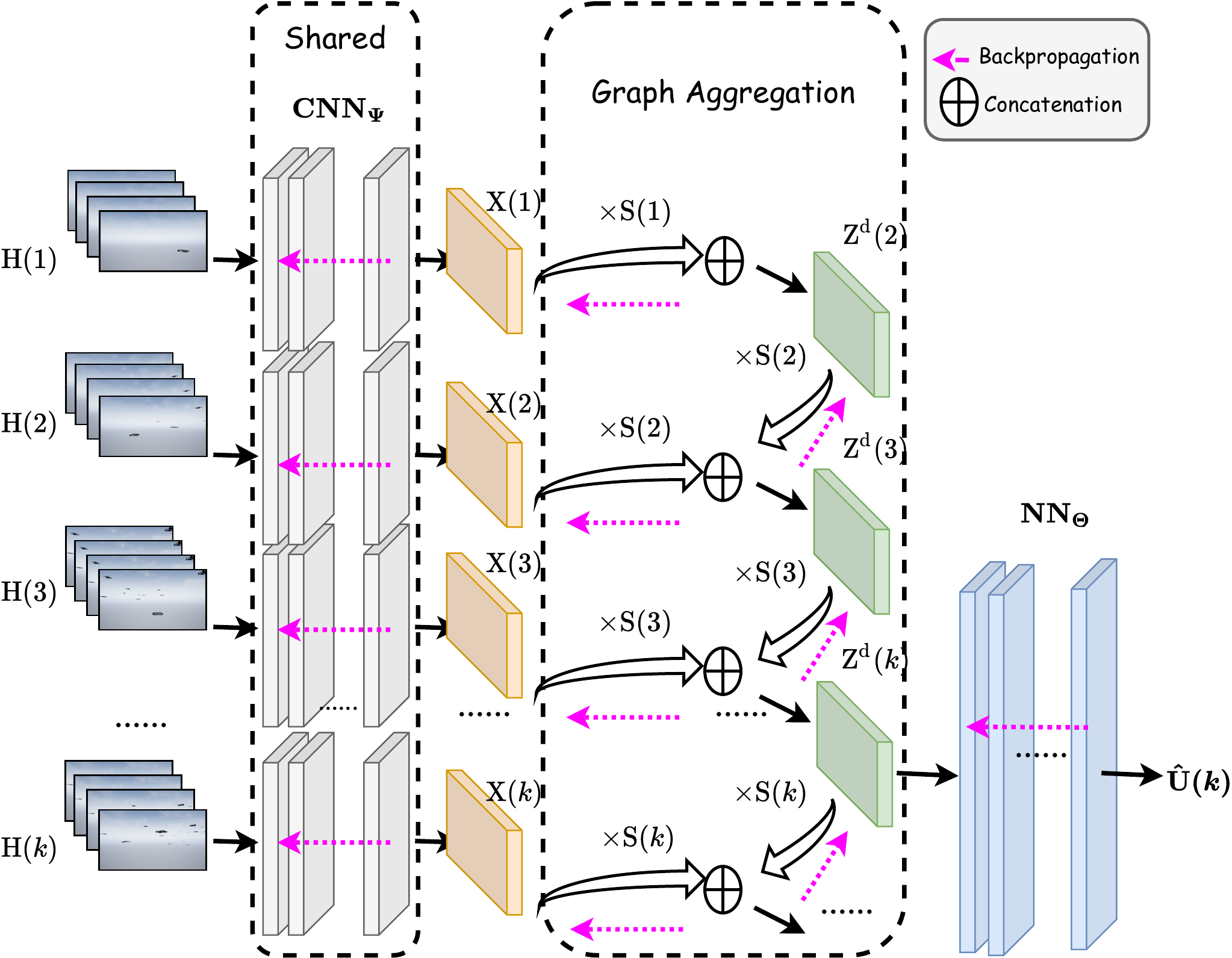}
    }
    \subfloat[GRNN\label{arch:GRNN}]{
        \centering
        \includegraphics[height=0.29\textheight]{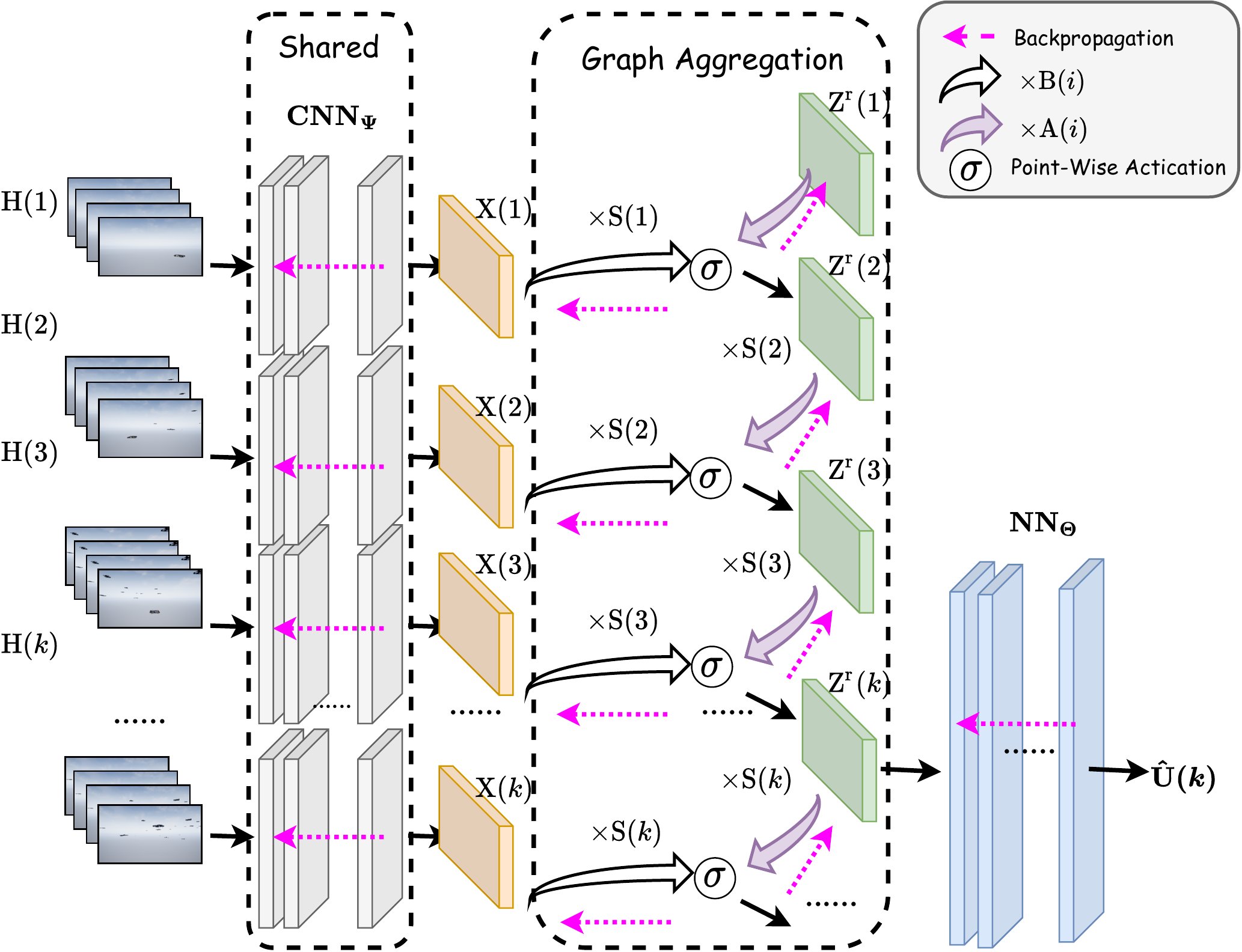}
    }
    \caption{Overview of the end-to-end training of VGAI framework. We examine two types of graph neural networks, DAGNN (Fig.~\ref{arch:DAGNN}) and GRNN (Fig.~\ref{arch:GRNN}). The collected neighboring information $\bbZ(t)$ comes from either graph sequence \eqref{eqn:aggregationSequence} or recurrent graph aggregation \eqref{eqn:GRNN}. We consecutively execute graph aggregation from time step $1$ to $K$ (\ie for $K-1$ steps). Then backward propagation through time is carried out by passing through  $\text{NN}_{\bbTheta}(\cdot)$ once and unrolling $\text{CNN}_{\bbPsi}(\cdot)$ $K$ times.}
    \label{fig:end2end}
\end{figure*}

Because the local graph aggregation \eqref{eqn:aggregationSequence} and graph convolutions \eqref{eqn:GRNN} are both fully differentiable, $\text{CNN}_{\bbPsi}(\cdot)$ can be jointly learned with $\text{NN}_{\bbTheta}(\cdot)$, by end-to-end backward propagation through time (BPTT), as illustrated in Figure~\ref{fig:end2end}. We update the parameters $\{\bbPsi, \bbTheta\}$ by $K$-step BPTT, given the $K$ consecutive GSO matrices $\{\bbS(t)\}_{t=1}^{K}$ and images $\{\bbH(t)\}_{t=1}^{K}$.

\blue{We emphasize that imitation learning renders the training stage a centralized one.} This is the case, not only because of the availability of an expert controller that is centralized, but also because of the weight-sharing scheme imposed by the CNN and the DAGNN or the GRNN. That is, the filter taps learned during the CNN/GNN stage are the same across all nodes [cf. \eqref{eqn:NN}, \eqref{eqn:graph-conv}]. This weight-sharing scheme is what allows the resulting architecture to scale and transfer successfully \cite{Ruiz2021-GNNs, Gama2020-Stability, Pfrommer2021-Discriminability}, as is later shown in the experiments (Section~\ref{sec:sims}). In any case, once the training phase is over, the resulting controller is, in fact, totally decentralized, as each filter tap can be stored at each agent separately, and thus each agent is capable of successfully computing the corresponding controller.

%As describe in Section \ref{sec:visual}, both $\text{CNN}_{\bbPsi}(\cdot)$ and $\text{NN}_{\bbTheta}(\cdot)$ are shared to each individual agent due to (1) the scalability to large swarms and (2) robustness to single-point failures during inference. Note that each training batch is randomly sampled from the training trajectories $\ccalT$ with different temporal indices $t$ and spatial indices $i$, so that the training of VGAI would not be biased on temporal or spatial dependencies, improving the robustness during dencentralized inference.

%%%%%%%%%%%%%%%%%%%%%%%%%%%%%%%%%%%%%%%%
%%%%             FIGURE             %%%%  fig:end2end
%%%%%%%%%%%%%%%%%%%%%%%%%%%%%%%%%%%%%%%%
%% arch:DAGNN, arch:GRNN
%%
%%%%          End of FIGURE         %%%%
%%%%%%%%%%%%%%%%%%%%%%%%%%%%%%%%%%%%%%%%

%%%%%%%%%%%%%%%%%%%%%%%%%%%%%%%%%%%%%%%%%%%%%%%%%%%%%%%%%%%%%%%%%%%%%%%%%%%%%%%%
%%%%                                                                        %%%%
%%%%                          EXPERIMENTAL RESULTS                          %%%%
%%%%                                                                        %%%%
%%%%%%%%%%%%%%%%%%%%%%%%%%%%%%%%%%%%%%%%%%%%%%%%%%%%%%%%%%%%%%%%%%%%%%%%%%%%%%%%

\section{Experimental Results} \label{sec:sims}

%!TEX root = 00-VGAI.tex

%%%%%%%%%%%%%%%%%%%%%%%%%%%%%%%%%%%%%%%%%%%%%%%%%%%%%%%%%%%%%%%%%%%%%%%%%%%%%%%%
%%%%                                                                        %%%%
%%%%                          EXPERIMENTAL RESULTS                          %%%%
%%%%                                                                        %%%%
%%%%%%%%%%%%%%%%%%%%%%%%%%%%%%%%%%%%%%%%%%%%%%%%%%%%%%%%%%%%%%%%%%%%%%%%%%%%%%%%
%%%% sec:sims
%%%%%%%%%%%%%

In what follows, we present the experimental setup, results, and discussions for controlling a flock of robots using VGAI. The code is available at \url{http://github.com/VITA-Group/VGAI}.

%%%%%%%%%%%%%%%%%%%%%%%%%%%%%%%%%%%%%%%%%%%%%%%%%%%%%%%%%%%%%%%%%%%%%%%%%%%%%%%%
%%%%                    SUBSECTION : Experimental setup                     %%%%
%%%%%%%%%%%%%%%%%%%%%%%%%%%%%%%%%%%%%%%%%%%%%%%%%%%%%%%%%%%%%%%%%%%%%%%%%%%%%%%%
%%%% subsec:setup
%%%%%%%%%%%%

\subsection{Experimental setup} \label{subsec:setup}

We simulate several controllers using the flocking objective in equation \eqref{eqn:flockingObjective} as the evaluation measure. 
The controllers proposed in this paper, denoted as VGAI (CNN + DAGNN) and VGAI (CNN + GRNN) are thoroughly compared with the following ones, listed in Table \ref{tab:controllers}.

\begin{itemize}
    \item The {\em centralized controller} in \eqref{eqn:optimalSolution}. This provides an empirical lower bound, because the decentralized approaches seek to imitate this centralized solution. Therefore, in the following the reported velocity variation costs from equation 
    \eqref{eqn:flockingObjective} are normalized to that of the centralized controller.
    \item DAGNN and GRNN: The learning based DAGNN and GRNN controllers supplied with {\em known neighbor position and velocity features}. We follow the  implementations in \cite{Tolstaya2019-Flocking, Ruiz2020-GRNN}, such that each individual agent can measure and receive accurate position and velocity of its neighbors, and the decisions are made based on the non-linear aggregation of the received positions and velocities. This offers another empirical lower bound on VGAI's expected performance, since it acts directly on the exact knowledge of the relative position and velocity of neighbors, whereas VGAI extracts information from visual inputs. We note that it has been shown in \cite{Gama2021-ControlGNN} that DAGNN and GRNN outperform popular GCNN architectures \cite{Gama2020-GNNs} in a position-based flocking problem.
    \item Position-based: We include comparison with another position-based algorithm that does not rely on GNNs, as described in \cite{Tanner2003-Stable}. This controller assumes perfect knowledge of the state, and uses only one-hop communications. Therefore, this controller offers a benchmark for comparison with multi-hop information passing and visual input state inference.
\end{itemize}

%%%%%%%%%%%%%%%%%%%%%%%%%%%%%%%%%%%%%%%%
%%%%             TABLE              %%%%  tab:controllers
%%%%%%%%%%%%%%%%%%%%%%%%%%%%%%%%%%%%%%%%
%%
\begin{table}[!t]
    \centering
    \caption{Experimental controllers.}
    \label{tab:controllers}
    \begin{tabular}{l|l}
        Controller & Description   \\ \hline
        Centralized \eqref{eqn:optimalSolution} & Expert used for imitation \\ & learning \\ 
        Position-based \cite{Tanner2003-Stable} & Distributed controller \\ & with perfect neighbor state \\ 
        DAGNN \eqref{eqn:NN} / GRNN \eqref{eqn:GRNNoutput} & Graph NN controllers with \\ & perfect neighbor state \\ 
        VGAI (CNN \eqref{eqn:CNN} + DAGNN) & DAGNN with visual \\ & state processing \\
        VGAI (CNN + GRNN) & GRNN with visual \\ & state processing \\
    \end{tabular}
\end{table}
%%
%%%%          End of TABLE          %%%%
%%%%%%%%%%%%%%%%%%%%%%%%%%%%%%%%%%%%%%%%

%%%%%%%%%%%%%%%%%%%%%%%%%%%%%%%%%%%%%%%%
%%%%             TABLE              %%%%  tab:dim-k
%%%%%%%%%%%%%%%%%%%%%%%%%%%%%%%%%%%%%%%%
%% {tab:dim-k-disk, tab:dim-k-knn}
\begin{table*}[!t]
    \centering
    \caption{Algorithm performance (normalized cost) with increasing communications exchanges ($K-1$). (a) Disk communication model, and (b) KNN communication model. Cost is normalized to the centralized controller in \eqref{eqn:optimalSolution}. The position-based controller does not depend on $K$ and yielded a cost of $4.13$.}
    \label{tab:dim-k}
    \subfloat[Disk model\label{tab:dim-k-disk}]{
        \centering
        \begin{tabular}{c|c|c|c}
            \toprule
            $K-1$   & $1$   & $2$  & $3$  \\ \midrule
            DAGNN & $2.14$ & $1.82$   & $1.70$   \\ \midrule
            GRNN  & $1.99$ & $1.79$   & $1.66$   \\ \midrule
            VGAI (CNN + DAGNN)  & $2.67$ & $2.46$ & $2.26$   \\ \midrule
            VGAI (CNN + GRNN)  & $2.55$ & $2.33$   & $2.24$  \\ \bottomrule
        \end{tabular}
    }
    \hspace{0.25\columnwidth}
    \subfloat[KNN model\label{tab:dim-k-knn}]{
        \centering
        \begin{tabular}{c|c|c|c}
            \toprule
            $K-1$   & $1$   & $2$  & $3$  \\ \midrule
            DAGNN & $2.15$ & $1.88$   & $1.61$   \\ \midrule
            GRNN  & $2.13$ & $1.78$   & $1.55$   \\ \midrule
            VGAI (CNN + DAGNN)  & $2.66$ & $2.40$   & $2.24$   \\ \midrule
            VGAI (CNN + GRNN)  & $2.46$ & $2.33$ & $2.21$ \\ \bottomrule
        \end{tabular}
    }
\end{table*}
%%
%%%%          End of TABLE          %%%%
%%%%%%%%%%%%%%%%%%%%%%%%%%%%%%%%%%%%%%%%

%%%%%%%%%%%%%%%%%%%%%%%%%%%%%%%%%%%%%%%%
%%%%             TABLE              %%%%  tab:dim-f
%%%%%%%%%%%%%%%%%%%%%%%%%%%%%%%%%%%%%%%%
%% {tab:dim-f-disk, tab:dim-f-knn}
\begin{table*}[!t]
    \centering
    \caption{Change in normalized cost with feature size $F$, under (a) the Disk model, and (b) the K-NN model. The position-based controller is not dependent on $F$ and has cost $3.97$.}
    \label{tab:dim-f}
    \subfloat[Disk model\label{tab:dim-f-disk}]{
        \centering
        \begin{tabular}{c|c|c|c}
            \toprule
            $F$   & $6$   & $12$  & $24$  \\ \midrule
            DAGNN & $1.70$ & ---   & ---   \\ \midrule
            GRNN  & $1.66$ & ---   & ---   \\ \midrule
            VGAI (CNN + DAGNN)  & $2.37$ & $2.38$   & $2.26$   \\ \midrule
            VGAI (CNN + GRNN)  & $2.31$ & $2.26$ & $2.24$  \\ \bottomrule
            
        \end{tabular}
    }
    \hspace{0.25\columnwidth}
    \subfloat[KNN model\label{tab:dim-f-knn}]{
        \centering
        \begin{tabular}{c|c|c|c}
            \toprule
            $F$   & $6$   & $12$  & $24$  \\ \midrule
            DAGNN & $1.61$ & ---   & ---   \\ \midrule
            GRNN  & $1.55$ & ---   & ---   \\ \midrule
            VGAI (CNN + DAGNN)  & $2.29$ & $2.26$ & $2.24$   \\ \midrule
            VGAI (CNN + GRNN)  & $2.23$ & $2.23$   & $2.21$ \\ \bottomrule
            
        \end{tabular}            
    }
\end{table*}
%%
%%%%          End of TABLE          %%%%
%%%%%%%%%%%%%%%%%%%%%%%%%%%%%%%%%%%%%%%%

Each agent is equipped with a communication transceiver capable of processing messages every $T_{s}$ seconds. We model communications between agents by considering both a disk model, which assumes agents located within a radius $R$ can communicate perfectly; and a $K$-nearest neighbor (KNN) model, which assumes that the closest $K_{\text{NN}}$ agents can communicate perfectly. The default disk radius is $R = 1.5\text{m}$ for the former, and the default number of connected neighbors is $K_{\text{NN}}=10$, for the latter.

The baseline scenario considers a large group of $N = 50$ agents and a discretization time period  $T_{s} = 0.01\text{s}$. We run all experiments for $100$ time steps. Flocking divergence depends on the specific cases and initializations, but we generally observe that when the decentralized controller cost is about $3$ times that of the centralized controller then the agents fail to successfully flock, diverging in their individual trajectories.

The flock locations are initialized uniformly in a disc with radius $\sqrt{N}$, in order to normalize the density of agents with different flock sizes. Initial agent velocities are sampled uniformly from the interval $[-v_{\text{init}},+v_{\text{init}}]$ and then a bias for the whole flock is added,  sampled from $[-0.3\ v_{\text{init}},+0.3\  v_{\text{init}}]$. Unless otherwise specified, we use $v_{\text{init}} = 3.0 \text{m}/\text{s}$; see Section~\ref{subsec:initial}. To eliminate poor initial configurations, we reject cases when any agent fails to have at least two neighbors (under the disk communications model) or if any pair of agents are closer than $0.2 \text{m}$. 

We conducted our experiments using the Microsoft Airsim visualization environment, to render the visual inputs. Figure \ref{fig:vis-input} describes the camera configuration, and displays visual observation snapshot examples for one agent during testing. The parameterized neural networks in \eqref{eqn:NN} for local visual state estimation contain $4$ fully connected layers of $1024$ neurons and the ReLU activation function is adopted. The controller was trained to predict the best action class with an $L_{1}$ loss, and we used the Adam optimizer with a learning rate $0.001$ and forgetting factors of $0.9$ and $0.999$ \cite{Kingma2015-ADAM}. Acceleration controls are saturated at the range $[-30, 30] \text{m}/\text{s}^{2}$ to provide a physical performance limit, and this also improves the numerical stability of training. 

We adopt Data set Aggregation (DAGGer) \cite{Ross2011-NoRegret} for imitation learning, an iterative algorithm that augments data sets, proceeding as follows. At the first iteration, we use the optimal policy $\pi^{\star}$ to gather a data set $D^{(0)}$, and train a policy $\pi^{(1)}$ based on data set $D^{(0)}$. Then at iteration $n$, we use the policy $\hat{\pi} = \beta \pi^{\star} + (1-\beta) \pi^{(n-1)}$ to collect more trajectories $D^{(n)}$, and a policy $\pi^{(n)}$ is trained on data set $D = D^{(0)}\bigcup D^{(1)}... \bigcup D^{(n)}$. The final policy $\tilde{\pi}$ is trained on the augmented data set $D = D^{(0)}\bigcup D^{(1)}... \bigcup D^{(k)}$, where $k$ iterations are executed. 
In the experiments, we set $n=1$ and $\beta = 0.33$ to account for the inconsistency of training and testing phase state distributions. For the first phase, we start the simulation with $D^{(0)}$ containing $15$ trajectories, while for the second phase we augment $D$ by including $5$ more trajectories with controller $\hat{\pi}$.

We ran five sets of experiments, described below, and in each experiment we test using both the Disk and KNN connectivity models. Recall that algorithm performance is measured using the cost in \eqref{eqn:flockingObjective}, normalized to the cost of the centralized controller. Thus algorithm performance is always reported relative to the centralized controller that has a relative cost value of $1.0$. We note that the ability to flock is determined, not only by the evolution over time of the velocity variation --as measured by \eqref{eqn:flockingObjective}-- but also by the velocity variation at the last time instant. In this sense, we have observed empirically that total trajectory costs below $3.0$ imply that the swarm has successfully flocked by the last time instant.

%%%%%%%%%%%%%%%%%%%%%%%%%%%%%%%%%%%%%%%%%%%%%%%%%%%%%%%%%%%%%%%%%%%%%%%%%%%%%%%%
%%%%                 SUBSECTION : Hyperparameter selection                  %%%%
%%%%%%%%%%%%%%%%%%%%%%%%%%%%%%%%%%%%%%%%%%%%%%%%%%%%%%%%%%%%%%%%%%%%%%%%%%%%%%%%
%%%% subsec:hyperparams
%%%%%%%%%%%%

\subsection{Hyperparameter selection} \label{subsec:hyperparams}

\myparagraph{Number of Communication Exchanges $K-1$.} We set $F=24$, $v_{\text{init}}=3\text{m}/\text{s}$, and vary the number of communication exchanges given by $K-1$, that determines the depth of temporal information collection as described in \eqref{eqn:aggregationSequence}. Note that if $K-1=1$, only one-hop communication exchanges with neighbors occur, and there are no delayed communications. Table \ref{tab:dim-k} summarizes the results with various choices of $K-1$ under the two connectivity models (disk and KNN). 

In each case increasing the number of communication exchanges results in better performance (lower cost). First, we observe that all controllers perform better than the position-based controller (which has a cost of $4.13$, indicating that it fails to flock). Second, we note that the DAGNN based controllers generally perform worse than for GRNNs, implying that directly combining current states from the last hidden state (as GRNN does) is more effective than collecting delayed inputs (as DAGNN does).

\myparagraph{Visual State Feature Dimension $F$.} We set $K=4$ and 
vary the dimension size $F$ of the feature $\bbX(t)$ extracted from the visual estimation process. Note that increasing $F$ improves the representation power of the decentralized controller. However, this comes at the expense of communication bandwidth, since more values need to be transmitted to neighboring agents. 

Table \ref{tab:dim-f} shows performance with various choices of dimension $F$ for both networking models. Performance generally improves as the feature size grows. DAGNN and GRNN have the exact information with $F = 6$. In contrast, the VGAI methods rely on learned features, and have good performance even with $F$ as small as 6.

\medskip As a result of these two experiments, from now on we adopt $F=24$ and $K=4$ for the distributed controllers.

%%%%%%%%%%%%%%%%%%%%%%%%%%%%%%%%%%%%%%%%
%%%%             TABLE              %%%%  tab:dim-f
%%%%%%%%%%%%%%%%%%%%%%%%%%%%%%%%%%%%%%%%
%% {tab:dim-f-disk, tab:dim-f-knn}
\begin{table*}[!t]
    \centering
    \caption{Performance with different maximum initial velocity $v_{\text{init}}$, for (a) disk, and (b) KNN network models. Performance generally degrades with higher $v_{\text{init}}$.}
    \label{tab:init-v}
    \subfloat[Disk Model\label{tab:init-v-disk}]{
        \centering
        \begin{tabular}{c|c|c|c}
            \toprule
            $v_{\text{init}}$[$\text{m}/\text{s}$]   & $1$   & $2$  & $3$  \\ \midrule
            Position-based & $3.36$ & $3.57$   & $4.32$   \\ \midrule
            DAGNN & $1.58$ & $1.63$   & $1.70$   \\ \midrule
            GRNN  & $1.55$ & $1.61$   & $1.66$   \\ \midrule
            VGAI (CNN + DAGNN)  & $1.97$ & $2.10$ & $2.26$   \\ \midrule
            VGAI (CNN + GRNN)  & $1.98$  & $2.08$  & $2.24$   \\ \bottomrule
        \end{tabular}
    }
    \hspace{0.25\columnwidth}
    \subfloat[KNN model \label{tab:init-v-knn}]{
        \centering
        \begin{tabular}{c|c|c|c}
            \toprule
            $v_{\text{init}}$[$\text{m}/\text{s}$]   & $1$   & $2$  & $3$  \\ \midrule
            Position-based & $3.36$ & $3.57$ & $4.32$  \\ \midrule
            DAGNN & $1.54$ & $1.61$   & $1.61$   \\ \midrule
            GRNN  & $1.48$ & $1.52$   & $1.55$   \\ \midrule
            VGAI (CNN + DAGNN)  & $1.88$ & $2.11$ & $2.24$   \\ \midrule
            VGAI (CNN + GRNN)  & $1.86$ & $2.04$   & $2.21$  \\ \bottomrule
        \end{tabular}
    }
\end{table*}
%%
%%%%          End of TABLE          %%%%
%%%%%%%%%%%%%%%%%%%%%%%%%%%%%%%%%%%%%%%%

%%%%%%%%%%%%%%%%%%%%%%%%%%%%%%%%%%%%%%%%
%%%%             TABLE              %%%%  tab:init-R
%%%%%%%%%%%%%%%%%%%%%%%%%%%%%%%%%%%%%%%%
%%
\begin{table}[!t]
    \centering
    \caption{Performance under the disk connectivity model with increasing one-hop communication radius $R$. As $R$ grows, the proposed vision-based VGAI controllers approach the performance of the graph NN's with perfect knowledge of their neighbors state.}
    \label{tab:init-R}
    \begin{tabular}{c|c|c|c}
        \toprule
        $R$ [$\text{m}$]   & $1.0$   & $1.5$  & $2.0$  \\ \midrule
        Position-based & $4.58$  & $3.93$ & $2.57$ \\ \midrule
        DAGNN & $1.95$  & $1.70$  & $1.59$  \\ \midrule
        GRNN & $1.90$  & $1.61$  & $1.54$  \\ \midrule
        VGAI(CNN+DAGNN)  & $2.65$  & $2.26$  & $1.77$  \\ \midrule
        VGAI(CNN+GRNN)  & $2.51$  & $2.24$  & $1.67$  \\ \bottomrule
    \end{tabular}
\end{table}
%%
%%%%          End of TABLE          %%%%
%%%%%%%%%%%%%%%%%%%%%%%%%%%%%%%%%%%%%%%%

%%%%%%%%%%%%%%%%%%%%%%%%%%%%%%%%%%%%%%%%%%%%%%%%%%%%%%%%%%%%%%%%%%%%%%%%%%%%%%%%
%%%%             SUBSECTION : Initial conditions and networking             %%%%
%%%%%%%%%%%%%%%%%%%%%%%%%%%%%%%%%%%%%%%%%%%%%%%%%%%%%%%%%%%%%%%%%%%%%%%%%%%%%%%%
%%%% subsec:initial
%%%%%%%%%%%%

\subsection{Initial conditions and networking} \label{subsec:initial}

\myparagraph{Maximum Initial Velocity $v_{\text{init}}$.} Increasing the maximum initial velocity $v_{\text{init}}$ results in a more difficult flocking control problem, with potentially slower convergence and the possibility of diverging agents and loss of communications connectivity. We keep the communications update rate fixed, which limits the control update rate. To test this, we vary the maximum initial velocity $v_{\text{init}}$. 

The results are summarized in Table~\ref{tab:init-v}. In general, we observe that increasing $v_{\text{init}}$ makes the flocking task more challenging for all controllers. Note that both vision-based and state-based controllers maintain comparably stable behaviors with normalized costs close to $2$ for all $v_{\text{init}}$. However, the position-based controller, which is also a decentralized controller with perfect state knowledge, but using only one-hop communication, fails to successfully flock the agents, as evidenced by its cost being greater than $3$.

%%%%%%%%%%%%%%%%%%%%%%%%%%%%%%%%%%%%%%%%
%%%%             FIGURE             %%%%  fig:vis-input
%%%%%%%%%%%%%%%%%%%%%%%%%%%%%%%%%%%%%%%%
%% {env:default, env:real}
\begin{figure*}[!t]
    \centering
    \subfloat[\label{env:default}]{
        \includegraphics[width=.24\linewidth]{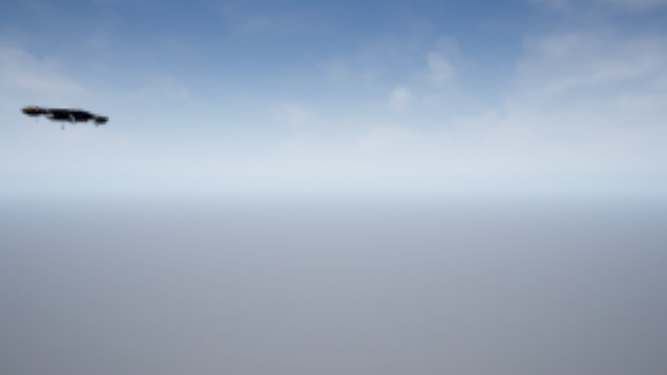}
        \includegraphics[width=.24\linewidth]{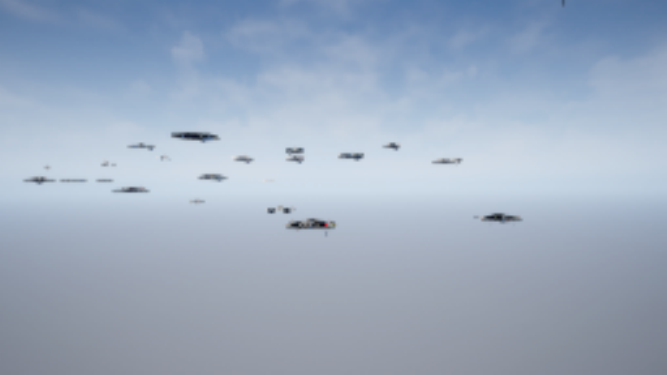}
        \includegraphics[width=.24\linewidth]{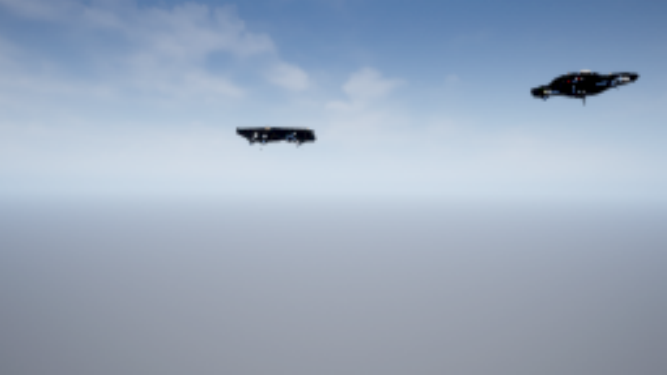}
        \includegraphics[width=.24\linewidth]{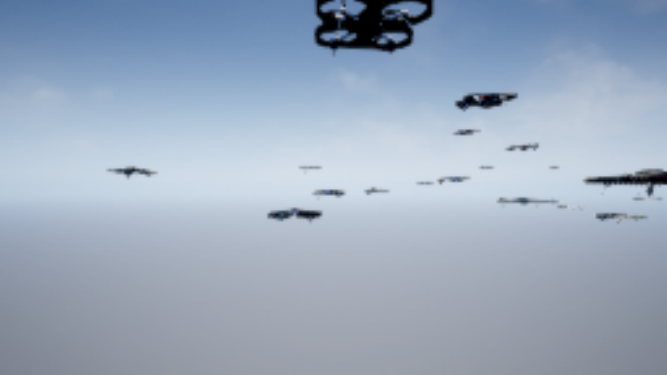}
    }
    \hfill
    \subfloat[\label{env:real}]{
        \includegraphics[width=.24\linewidth]{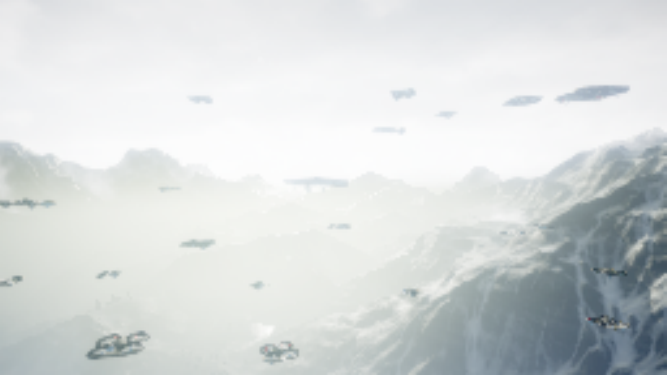}
        \includegraphics[width=.24\linewidth]{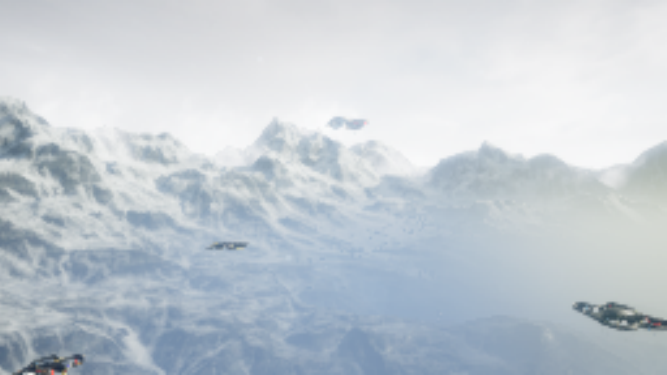}
        \includegraphics[width=.24\linewidth]{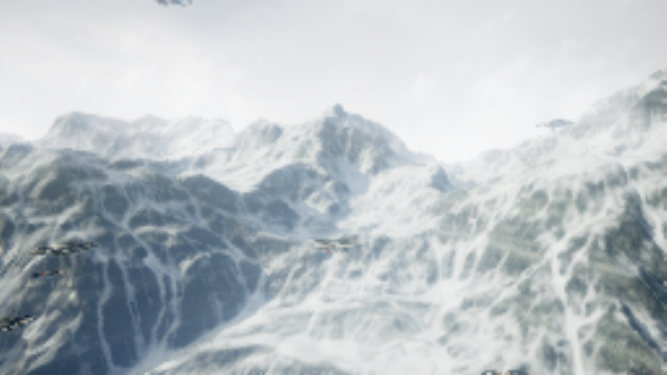}
        \includegraphics[width=.24\linewidth]{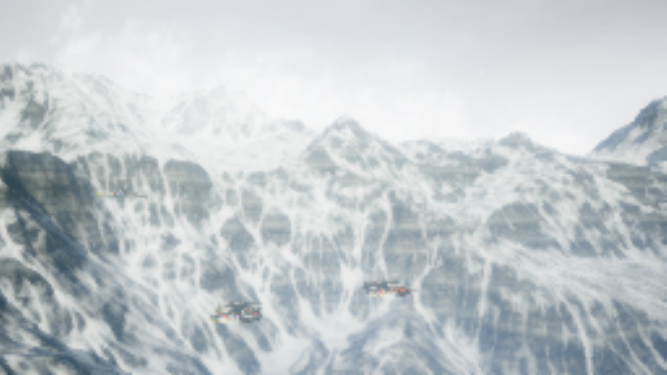}
    }
    \caption{Top panels~\ref{env:default} shows the Plain Sky rendered visual environment, and bottom panels~\ref{env:real} are from the Mountain Landscape environment. Experiments are conducted in the Plain Sky environment unless otherwise noted. Each agent has four $144 \times 256$ resolution cameras providing full visibility in the horizontal plane, pointing to front-center, front-left, front-right, and rear. The four views are concatenated for input to the visual processing neural network. For the camera configuration, we adjust the quaternion orientation of front-left and front-right cameras to $-\frac{\pi}{2}$ and $-\frac{\pi}{2}$ in order to maximally increase the visibility of each drone. We restrict random initial agent altitude to be in $[38, 41]$ m, so agents are within the camera viewing angles.  We also set the aptitude of each agent in-between $38\text{m}$ and $41\text{m}$ for collision-free random initialization. Airsim supports four cameras per drone, so our experiments are limited to 2D horizontal fields of view; additional cameras are needed for a 3D extension.
    %flocking and Note that our experiments target at 2D flocking problems, and each agent covers full field of view in the horizontal plane. For the 3D flocking problems, at least two cameras located at the top and bottom sides are expected to complete the 720-degree field of view. We leave it as our future works.}
    }
    \label{fig:vis-input}
\end{figure*}
%%
%%%%          End of FIGURE         %%%%
%%%%%%%%%%%%%%%%%%%%%%%%%%%%%%%%%%%%%%%%

%%%%%%%%%%%%%%%%%%%%%%%%%%%%%%%%%%%%%%%%
%%%%             TABLE              %%%%  tab:init-C
%%%%%%%%%%%%%%%%%%%%%%%%%%%%%%%%%%%%%%%%
%%
\begin{table}[!t]
    \centering
    \caption{Performance under the $K$ nearest neighbor connectivity model with increasing $K$. As $K$ grows, the proposed vision-based VGAI controllers approach the performance of the graph NN's with perfect knowledge of their neighbors state. }
        \label{tab:init-C}
    \begin{tabular}{c|c|c|c}
        \toprule
        $K_{\text{NN}}$    & $7$   & $10$  & $15$  \\ \midrule
        Position-based & $4.58$  & $3.93$ & $2.57$ \\ \midrule
        DAGNN & $1.82$  & $1.61$  & $1.49$  \\ \midrule
        GRNN & $1.79$  & $1.55$  & $1.40$  \\ \midrule
        VGAI(CNN+DAGNN)  & $2.49$  & $2.24$  & $1.73$  \\ \midrule
        VGAI(CNN+GRNN)  & $2.44$  & $2.21$  & $1.59$  \\ \bottomrule
    \end{tabular}
\end{table}
%%
%%%%          End of TABLE          %%%%
%%%%%%%%%%%%%%%%%%%%%%%%%%%%%%%%%%%%%%%%

\myparagraph{Networking Disk Model Radius $R$.} Under the disk model, only agents within $R$ meters communicate at any given time, so the node degree of each agent is time-varying. Increasing $R$ enables more one-hop connectivity and decreased delay with respect to more distant agents, and should result in better controller performance. We see in Table~\ref{tab:init-R} that this is, indeed, the case. We note that as $R$ increases the performance gap between perfect state information in DAGNN and GRNN, and our proposed vision-based VGAI  controllers, shrinks considerably.

\myparagraph{Number of Connected Neighbors in the KNN Model.} In the KNN model, each agent connects to the $K_{\text{NN}}$ nearest neighbors, and we expect that controller performance will be enhanced as increasing $K_{\text{NN}}$ results in a more densely connected network topology. This is demonstrated in the results of Table~\ref{tab:init-C}. Similar to the disk model, increasing the one-hop connectivity results in better performance, and the gain is more noticeable in the VGAI controllers as they exchange visual features. 

%%%%%%%%%%%%%%%%%%%%%%%%%%%%%%%%%%%%%%%%%%%%%%%%%%%%%%%%%%%%%%%%%%%%%%%%%%%%%%%%
%%%%              SUBSECTION : Generalization to larger teams               %%%%
%%%%%%%%%%%%%%%%%%%%%%%%%%%%%%%%%%%%%%%%%%%%%%%%%%%%%%%%%%%%%%%%%%%%%%%%%%%%%%%%
%%%% subsec:generalization
%%%%%%%%%%%%

\subsection{Generalization to larger teams} \label{subsec:generalization}

%%%%%%%%%%%%%%%%%%%%%%%%%%%%%%%%%%%%%%%%
%%%%             TABLE              %%%%  tab:generality
%%%%%%%%%%%%%%%%%%%%%%%%%%%%%%%%%%%%%%%%
%%
\begin{table}[!t]
    \centering
    \caption{Generalization to larger teams: flocking performance versus team size. The VGAI controller was trained with $N=50$ agents, and then tested with size $N'$ teams.}
    \label{tab:generality}
    \begin{tabular}{c|c|c|c}
        \toprule
        $N'$ & 50    & 60    & 75    \\ \midrule
        VGAI(CNN+DAGNN) & $2.24$ & $2.28$ & $2.41$ \\ \midrule
        VGAI(CNN+GRNN) & $2.26$ & $2.42$ & $2.66$ \\ \bottomrule
    \end{tabular}
\end{table}
%%
%%%%          End of TABLE          %%%%
%%%%%%%%%%%%%%%%%%%%%%%%%%%%%%%%%%%%%%%%

Scaling up the number of agents is a critical issue for decentralized control problems. Recall that, while our approach relies on communicating with neighbors [cf. \eqref{eqn:aggregationSequence} and  \eqref{eqn:GRNN}], we learn a single controller that is then used by all agents [cf. \eqref{eqn:NN}]. We hypothesize that, due to the permutation equivariance and stability properties of GNNs \cite{Gama2020-Stability, Ruiz2020-GRNN}, we can scale up the number of agents after training. This is a particularly critical property in the context of imitation learning, whereby oftentimes we rely on an expert controller that is centralized and difficult to scale. Therefore, the ability to learn distributed controllers in smaller team offline settings, and then being able to deploy these learned controllers in larger teams online at execution time, is of paramount importance.

To test this, we trained with $N=50$ agents, and then applied this controller to $N' = 60$ and $75$ agents. The results are shown in Table~\ref{tab:generality}. Most notably, the VGAI controller is capable of successfully flocking these teams (cost less than $3$) and, in fact, it exhibits only a relatively small cost increase, even when scaling up the size of the team by $50\%$. 

%%%%%%%%%%%%%%%%%%%%%%%%%%%%%%%%%%%%%%%%%%%%%%%%%%%%%%%%%%%%%%%%%%%%%%%%%%%%%%%%
%%%%            SUBSECTION : Performance with visual degradation            %%%%
%%%%%%%%%%%%%%%%%%%%%%%%%%%%%%%%%%%%%%%%%%%%%%%%%%%%%%%%%%%%%%%%%%%%%%%%%%%%%%%%
%%%% subsec:vis-deg
%%%%%%%%%%%%

\subsection{Performance with visual degradation} \label{subsec:vis-deg}

\begin{figure}[!t]
    \centering
    \subfloat[\label{deg:default}]{
        \includegraphics[width=.31\linewidth]{figures/front.pdf}
        \includegraphics[width=.31\linewidth]{figures/left.pdf}
        \includegraphics[width=.31\linewidth]{figures/right.pdf}
    }
    \hfill
    \subfloat[\label{deg:gaussian}]{
        \includegraphics[width=.31\linewidth]{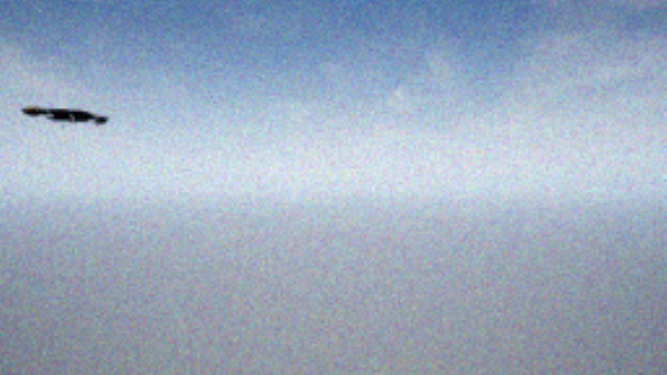}
        \includegraphics[width=.31\linewidth]{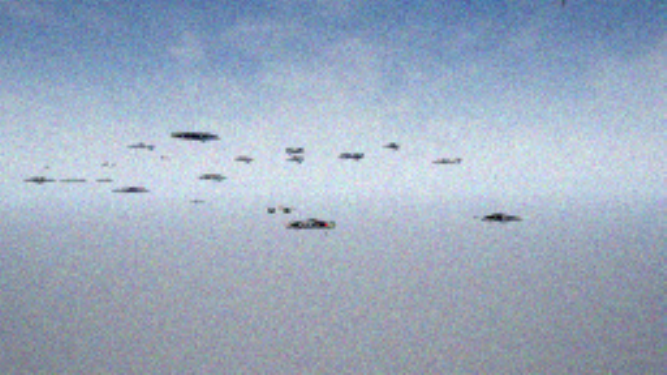}
        \includegraphics[width=.31\linewidth]{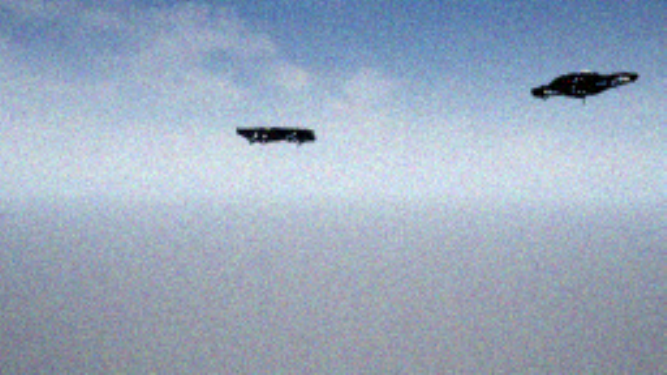}
    }
    \hfill
    \subfloat[\label{deg:blurring}]{
        \includegraphics[width=.31\linewidth]{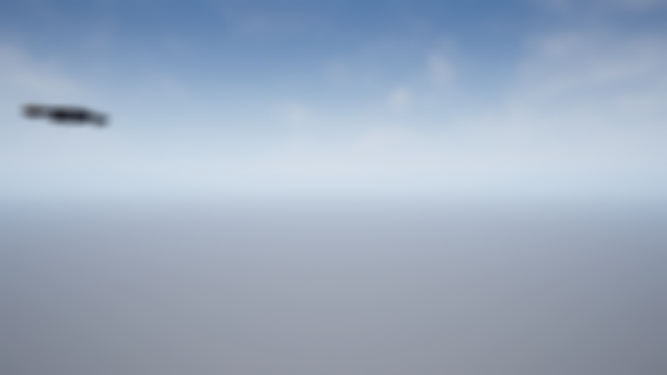}
        \includegraphics[width=.31\linewidth]{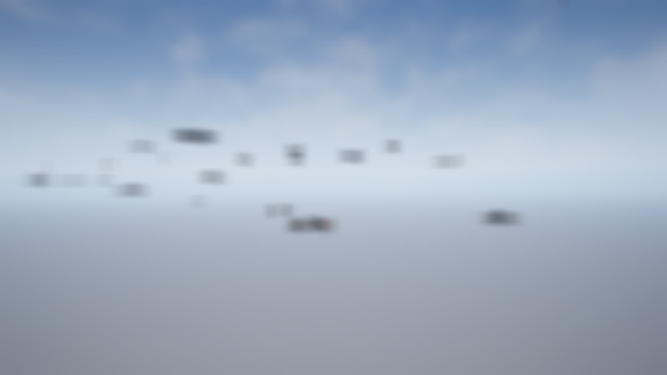}
        \includegraphics[width=.31\linewidth]{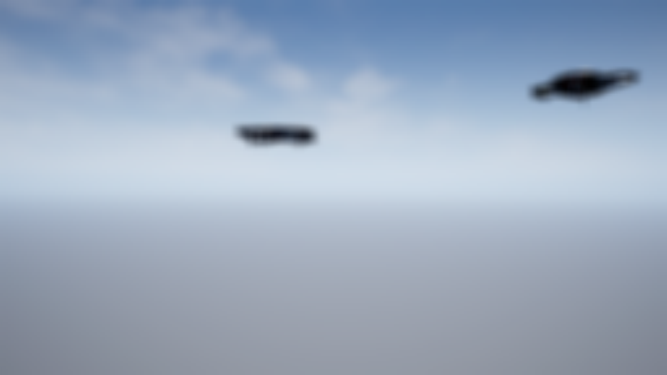}
    }
    \caption{Example original and degraded input images. Figure~\ref{deg:default}: original rendered images,  Figure~\ref{deg:gaussian}: additive high variance white Gaussian noise, Figure~\ref{deg:blurring}: blurring.}
    \label{fig:vis-deg}
\end{figure}

The experiments described so far were visually rendered with the sky scene, illustrated in Figure~\ref{env:default}. While training may be conducted under good conditions, fielded systems may suffer some visual input degradation from a variety of sources. In this experiment we consider the VGAI controller trained offline in visually favorable scenarios but tested online in challenging visual degradation cases, adding very high variance white Gaussian noise with distribution $N(0,100)$ and blurring with a $5 \times 5$ unity valued kernel. Examples of degraded images are shown in Figure~\ref{fig:vis-deg}. This tests the ability of VGAI to extract useful state information despite significant visual loss compared to the training data.

Results in Table \ref{tab:vis-generality} show relatively small performance loss (increased cost) for both additive noise and blurring. While successful flocking was achieved with both forms of visual degradation, blurring resulted in higher cost than adding very high variance white noise. This seems reasonable because the positioning of neighbors is visually conveyed through relative size, and blurring may degrade edge location information more than spatially uncorrelated noise.

%%%%%%%%%%%%%%%%%%%%%%%%%%%%%%%%%%%%%%%%
%%%%             TABLE              %%%%  tab:vis-generality
%%%%%%%%%%%%%%%%%%%%%%%%%%%%%%%%%%%%%%%%
%%
\begin{table}[!t]
    \centering
    \caption{System performance with visual degradation. Training used unaltered images, and test images had high variance additive white Gaussian noise or blurring. Flocking was successful in both cases.}
    \label{tab:vis-generality}
    \begin{tabular}{c|c|c|c}
        \toprule
        Input & Default & Gaussian & Blurring    \\ \midrule
        VGAI(CNN+DAGNN) & $2.26$ & $2.36$ & $2.54$  \\ \midrule
        VGAI(CNN+GRNN) & $2.24$ & $2.33$ & $2.61$  \\ \bottomrule
    \end{tabular}
\end{table}
%%
%%%%          End of TABLE          %%%%
%%%%%%%%%%%%%%%%%%%%%%%%%%%%%%%%%%%%%%%%

%%%%%%%%%%%%%%%%%%%%%%%%%%%%%%%%%%%%%%%%%%%%%%%%%%%%%%%%%%%%%%%%%%%%%%%%%%%%%%%%
%%%%                SUBSECTION : Enhancing visual processing                %%%%
%%%%%%%%%%%%%%%%%%%%%%%%%%%%%%%%%%%%%%%%%%%%%%%%%%%%%%%%%%%%%%%%%%%%%%%%%%%%%%%%
%%%% subsec:vis-generalization
%%%%%%%%%%%%

%%%%%%%%%%%%%%%%%%%%%%%%%%%%%%%%%%%%%%%%
%%%%             FIGURE             %%%%  fig:vis-deg
%%%%%%%%%%%%%%%%%%%%%%%%%%%%%%%%%%%%%%%%
%% {deg:default, deg:gaussian, deg:blurring}
%%
%%%%          End of FIGURE         %%%%
%%%%%%%%%%%%%%%%%%%%%%%%%%%%%%%%%%%%%%%%

\subsection{Enhancing visual processing} \label{subsec:vis-generalization}

The VGAI controller is modularized, making it straightforward to change the simple vision processing CNN (Fig.~\ref{fig:uni_droNet}) with other visual processing architectures, including off-the-shelf alternatives. For example, enhanced visual processing can provide semantic grouping and object localization abilities, to gain agent location information for navigating in complex environments. To illustrate this, we replace our default DroNet with either a VGG neural network \cite{Simonyan2015-VGG} or a pre-trained object detector Yolo-V3 \cite{Redmon2018-YOLOv3}. The VGG neural network consists of $8$ plain convolutional kernels and $3$ fully-connected layers, serving as another comparison variant of DroNet.  The pre-training of Yolo-V3 uses bounding-box annotation for the observed agents in the scene; refer to Appendix~\ref{subsec:yolo-detail} for details. The VGG or the pre-trained Yolo-V3 neural network will serve as the visual state estimator of the VGAI, while all other parts in the end-to-end pipeline, \eg the graph aggregation and the action inference network, remain the same [cf. Fig~\ref{fig:VGAI}].

To test the increased power of this newly adopted visual state estimator, we consider a plain sky scenario as well as a mountain landscape environment; see Fig.~\ref{fig:vis-input}. Results are shown in Table~\ref{tab:vis-real}. We observe that the flocking behavior converged more slowly in the mountain landscape environment, while still achieving successful flocking, 
thanks to employing the pre-trained Yolo-V3 object detector.
%\green{In short, adding a pre-trained Yolo-V3 object detector, trained with pre-annotated images seems to be necessary for convergence in more complex settings. In any case, } 
The results highlight the adaptability of the VGAI architecture, enabling more sophisticated visual front-end processing and application in different complex environments.

%%%%%%%%%%%%%%%%%%%%%%%%%%%%%%%%%%%%%%%%
%%%%             TABLE              %%%%  tab:vis-real
%%%%%%%%%%%%%%%%%%%%%%%%%%%%%%%%%%%%%%%%
%%
\begin{table}[!t]
    \centering
    \caption{Change in normalized cost with a more complex visual environment. The default CNN, dubbed DroNet, was replaced with either a VGG NN or a pre-trained Yolo-V3 NN for the Mountain Landscape. The position-based controller yielded a cost of $3.97$.}
    \label{tab:vis-real}
    \begin{tabular}{c|c|c}
        \toprule
        Environment & Plain Sky & Mountain Landscape \\ \midrule
        VGAI(DroNet+DAGNN) & $2.26$ & $> 4$  \\ \midrule
        VGAI(DroNet+GRNN) & $2.24$ & $> 4$  \\ \bottomrule
        VGAI(VGG+DAGNN) & $2.29$ & $ > 4$  \\ \midrule
        VGAI(VGG+GRNN) & $2.23$ &  $> 4$  \\ \bottomrule
        VGAI(Yolo-V3+DAGNN) & $2.13$ & $2.89$  \\ \midrule
        VGAI(Yolo-V3+GRNN) & $2.07$ & $2.77$  \\ \bottomrule
    \end{tabular}
\end{table}
%%
%%%%          End of TABLE          %%%%
%%%%%%%%%%%%%%%%%%%%%%%%%%%%%%%%%%%%%%%%

%%%%%%%%%%%%%%%%%%%%%%%%%%%%%%%%%%%%%%%%%%%%%%%%%%%%%%%%%%%%%%%%%%%%%%%%%%%%%%%%
%%%%                        SUBSECTION : Discussion                         %%%%
%%%%%%%%%%%%%%%%%%%%%%%%%%%%%%%%%%%%%%%%%%%%%%%%%%%%%%%%%%%%%%%%%%%%%%%%%%%%%%%%
%%%% subsec:discussion
%%%%%%%%%%%%

\subsection{Discussion} \label{subsec:discussion}

In general, we observe that the VGAI controller consistently manages to successfully flock the robot swarm in all our experiments. We also observe that its performance surpasses the position-based controller \cite{Tanner2003-Stable} for all experiments. As expected, the centralized expert controller \eqref{eqn:optimalSolution} acts as an upper bound on the performance of VGAI, at the expense of access to perfect global information. The DAGNN and GRNN controllers acting on perfect knowledge of the state of the $K$-hop neighbors also act as lower bounds on the performance of VGAI. Nonetheless, VGAI provides a successful vision-based distributed controller with performance oftentimes matching decentralized controllers with perfect knowledge of local information.

\begin{figure}[!t]
    \centering
    \subfloat[\label{fig:local-trajetory}]{
        \includegraphics[width=0.8\linewidth]{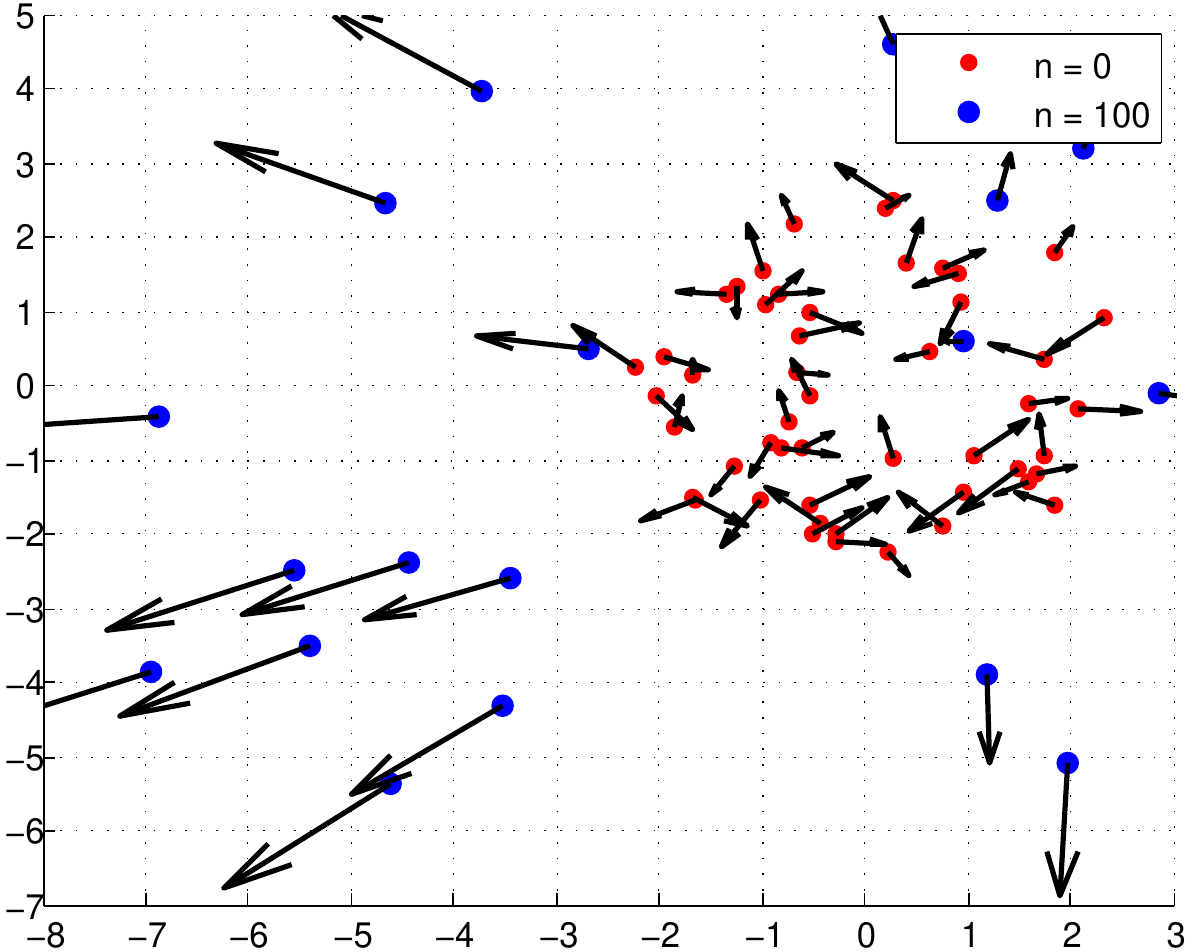}
    }
    \hfill
    \subfloat[\label{fig:vgai-trajetory}]{
        \includegraphics[width=0.8\linewidth]{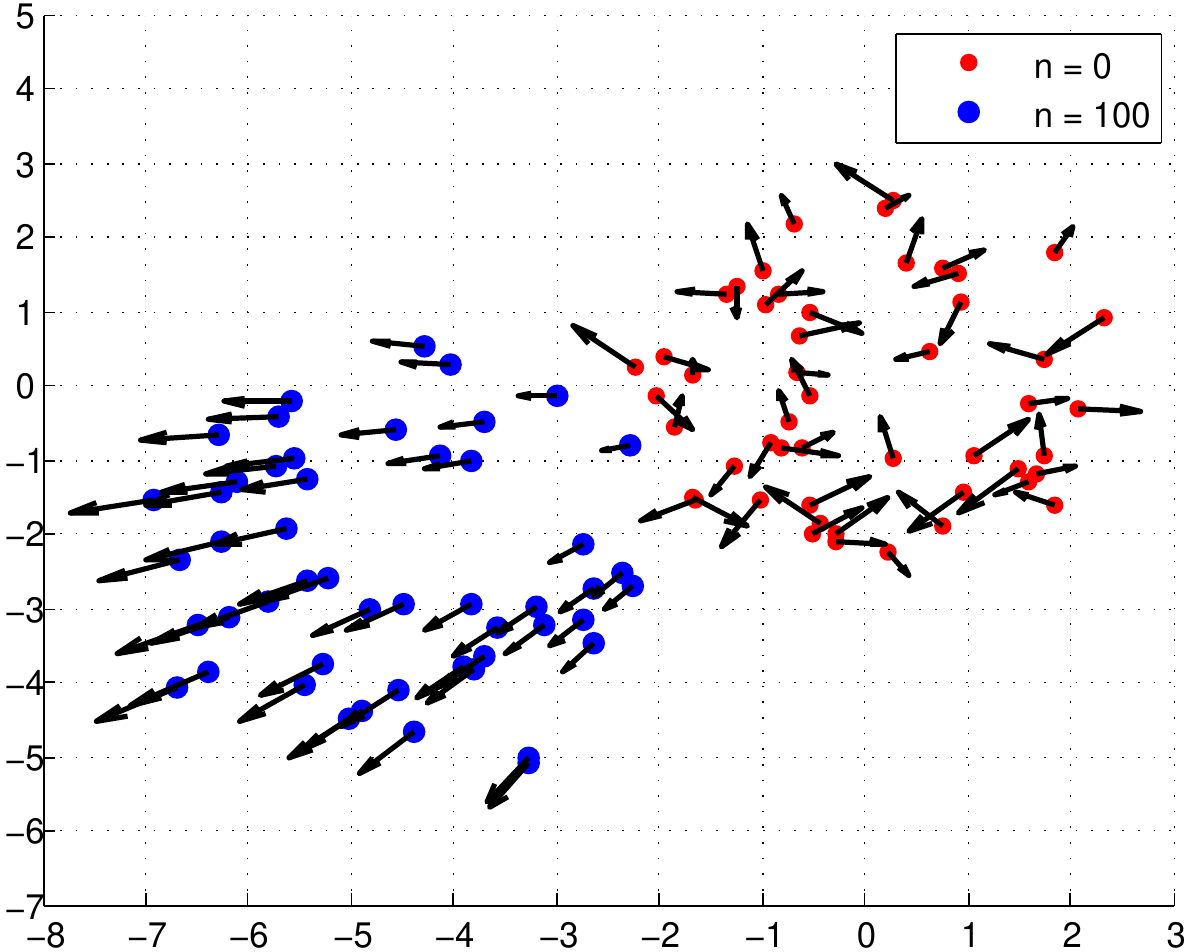}
    }
    \caption{ Flocking examples at initial $(n=0)$ and final $(n=100)$ steps, for the position-based controller \ref{fig:local-trajetory}, and VGAI controller \ref{fig:vgai-trajetory}. The position-based controller diverges, whereas the VGAI controller converges and maintains agent spacing. }
    %The two plots \ref{fig:local-trajetory} and \ref{fig:vgai-trajetory} illustrate the flock positions controlled by local controller and VGAI controller, respectively. VGAI controller produces the desired consensus in agents’ velocities and regular spacing between agents, while the local controller makes the flock scatter.}
    \label{fig:trajetory}
\end{figure}

To qualitatively analyze the flocking qualitity of the VGAI we compare it with trajectories observed by using the position-based controller \cite{Tanner2003-Stable}. Fig. \ref{fig:trajetory} illustrates the trajectory evolution of a VGAI controller (CNN+GRNN) versus the position-based one. Each sub-figure shows the initial agent positions and velocities at different time instants, qualitatively illustrating the trajectories of the agents, where it is observed that VGAI successfully flocks the swarm, while the position-based controller fails to do so.

Finally, we note that while a physical implementation of the VGAI controller is outside of the scope of this work, the ability of the controller to adapt to complex, noisy images (see Sections~\ref{subsec:vis-deg}~and~\ref{subsec:vis-generalization}), holds promise for deployment.

%%%%%%%%%%%%%%%%%%%%%%%%%%%%%%%%%%%%%%%%
%%%%             FIGURE             %%%%  fig:trajetory
%%%%%%%%%%%%%%%%%%%%%%%%%%%%%%%%%%%%%%%%
%% {fig:local-trajetory, fig:vgai-trajetory}
%%
%%%%          End of FIGURE         %%%%
%%%%%%%%%%%%%%%%%%%%%%%%%%%%%%%%%%%%%%%%

%%%%%%%%%%%%%%%%%%%%%%%%%%%%%%%%%%%%%%%%%%%%%%%%%%%%%%%%%%%%%%%%%%%%%%%%%%%%%%%%
%%%%                                                                        %%%%
%%%%                              CONCLUSIONS                               %%%%
%%%%                                                                        %%%%
%%%%%%%%%%%%%%%%%%%%%%%%%%%%%%%%%%%%%%%%%%%%%%%%%%%%%%%%%%%%%%%%%%%%%%%%%%%%%%%%

\section{Conclusions} \label{sec:conclusions}

%!TEX root = 00-VGAI.tex

%%%%%%%%%%%%%%%%%%%%%%%%%%%%%%%%%%%%%%%%%%%%%%%%%%%%%%%%%%%%%%%%%%%%%%%%%%%%%%%%
%%%%                                                                        %%%%
%%%%                              CONCLUSIONS                               %%%%
%%%%                                                                        %%%%
%%%%%%%%%%%%%%%%%%%%%%%%%%%%%%%%%%%%%%%%%%%%%%%%%%%%%%%%%%%%%%%%%%%%%%%%%%%%%%%%
%%%% sec:conclusions
%%%%%%%%%%%%%%%%%%%%

We presented Vision-based Graph Aggregation and Inference (VGAI), a decentralized PAC-loop controller learning framework for large-scale robot swarms. We demonstrated the feasibility of a CNN-GNN network cascade for learning the local controller, using imitation learning based on a centralized control solution. This approach works with large teams of agents based only on local visual observation, with coupled state dynamics and sparse communication links. Experimental results quantitatively confirm the value of local neighborhood information for the stability of controlled flocks. We also showed that the VGAI framework is robust to changes in the communications graph topology, number of agents, and random velocity initialization. The method is robust to visual degradation, and the visual feature extraction neural network can be enhanced or replaced for complex visual environments without altering the VGAI architecture. 

This work opens up exciting avenues for future research. The extension to three-dimensional environments can be carried out, and the use of different visual feature extractors in combination with image segmentation algorithms can be used to enhance the performance and ability to work in complex visual environments. Other forms of perception might also be employed, perhaps in combination with vision, and other communication networking schemes can be devised and tested.  Implementation and testing of the VGAI controller on physical platforms is also of interest.   Finally, we note that the overall approach is applicable to other distributed control problems in general.

%%%%%%%%%%%%%%%%%%%%%%%%%%%%%%%%%%%%%%%%%%%%%%%%%%%%%%%%%%%%%%%%%%%%%%%%%%%%%%%%
%%%%                                                                        %%%%
%%%%                                APPENDIX                                %%%%
%%%%                                                                        %%%%
%%%%%%%%%%%%%%%%%%%%%%%%%%%%%%%%%%%%%%%%%%%%%%%%%%%%%%%%%%%%%%%%%%%%%%%%%%%%%%%%

% or



\appendices

%!TEX root = 00-VGAI.tex

%%%%%%%%%%%%%%%%%%%%%%%%%%%%%%%%%%%%%%%%%%%%%%%%%%%%%%%%%%%%%%%%%%%%%%%%%%%%%%%%
%%%%                                                                        %%%%
%%%%                   APPENDIX : IMPLEMENTATION DETAILS                    %%%%
%%%%                                                                        %%%%
%%%%%%%%%%%%%%%%%%%%%%%%%%%%%%%%%%%%%%%%%%%%%%%%%%%%%%%%%%%%%%%%%%%%%%%%%%%%%%%%
%%%% subsec:yolo-detail
%%%%%%%%%%%%%%%%%%%%%%%

\section{Object Detector Implementation Details} \label{subsec:yolo-detail}

In this Appendix we provide further details for the experiment in Section \ref{subsec:vis-generalization}, testing with more complex visual backgrounds. We employ an object detector based NN architecture for visual state estimation to improve robustness. We first train a Yolo detector \cite{Redmon2018-YOLOv3} on the background scene dataset. The resulting YOLO parameters are fixed thereafter. The raw features of the detected objects $\tilde{\bf X}(t) \in\mathbb{R}^{\tilde{f} \times {N_{obj}}}$ are obtained through the trained YOLO network for every image. Here ${N_{obj}}$ is the number of detected UAVs, and $\tilde{f}=5$ corresponds to the coordinates $\tilde{x_i}$, $\tilde{y_i}$, the height $\tilde{h_i}$, the width $\tilde{w_i}$ and the confidence score $\tilde{c_i}$ of the bounding box of each detected object (UAV). The visual features ${\bf X}(t)\in \mathbb{R}^9$ are then extracted from $\tilde{\bf X}(t)$ and are fed into the graph aggregation stage of VGAI. The visual features ${\bf X}(t)$ are a combination of $xy$-coordinates and the area for the nearest object, averaged top 3 nearest objects, and averaged all detected objects.  Denoting the confident area as $\tilde{s_i}=\tilde{w_i}\tilde{h_i}\tilde{c_i}$ for $i=1,2,\dots,{N_{obj}}$, then we have,
\begin{equation}
    [{\bf X}(t)]_{1,2,3} = [\tilde{x_0},\tilde{y_0},\tilde{s_0}],
\end{equation}
\begin{equation}
    [{\bf X}(t)]_{4,5,6} = \Big[\frac{\sum_{i=1}^3 \tilde{x_i}\tilde{s_i}}{\sum_{i=1}^3 \tilde{s_i}},\frac{\sum_{i=1}^3 \tilde{y_i}\tilde{s_i}}{\sum_{i=1}^3 \tilde{s_i}},\frac{\sum_{i=1}^3 \tilde{s_i}}{3}\Big],
\end{equation}
\begin{equation}
    [{\bf X}(t)]_{7,8,9} = \Big[\frac{\sum_{i=1}^{N_{obj}} \tilde{x_i}\tilde{s_i}}{\sum_{i=1}^{N_{obj}} \tilde{s_i}},\frac{\sum_{i=1}^{N_{obj}} \tilde{y_i}\tilde{s_i}}{\sum_{i=1}^{N_{obj}} \tilde{s_i}},\frac{\sum_{i=1}^{N_{obj}} \tilde{s_i}}{{N_{obj}}}\Big],
\end{equation}
where the tuples ($\tilde{x_i},\tilde{y_i},\tilde{s_i}$) are sorted according to $\tilde{s_i}$, the area. 

%%%%%%%%%%%%%%%%%%%%%%%%%%%%%%%%%%%%%%%%%%%%%%%%%%%%%%%%%%%%%%%%%%%%%%%%%%%%%%%%
%%%%                                                                        %%%%
%%%%                               REFERENCES                               %%%%
%%%%                                                                        %%%%
%%%%%%%%%%%%%%%%%%%%%%%%%%%%%%%%%%%%%%%%%%%%%%%%%%%%%%%%%%%%%%%%%%%%%%%%%%%%%%%%



\bibliographystyle{bibFiles/IEEEtranD}
\bibliography{bibFiles/myIEEEabrv,bibFiles/biblioVGAI}

\end{document}